\pdfoutput=1

\documentclass[11pt]{article}

\usepackage{ACL2023}

\usepackage{times}
\usepackage{latexsym}

\usepackage[T1]{fontenc}

\usepackage[utf8]{inputenc}

\usepackage{microtype}

\usepackage{inconsolata}
\usepackage{multirow}
\usepackage{amssymb}
\usepackage{graphicx}
\usepackage{color}
\usepackage{amsmath}
\usepackage{enumitem}
\usepackage{mdframed}
\usepackage{bm}
\usepackage{arydshln}
\usepackage{tcolorbox}
\usepackage{algorithm}
\usepackage{algorithmicx}
\usepackage[noend]{algpseudocode}
\usepackage{subcaption}
\usepackage[export]{adjustbox}
\usepackage{rotating}
\usepackage{booktabs}
\usepackage{marvosym}
\usepackage{footmisc}
\usepackage{amssymb}
\usepackage[utf8]{inputenc} 
\usepackage{colortbl}
\usepackage{pifont}

\title{Browse and Concentrate: \\ Comprehending Multimodal Content via prior-LLM Context Fusion}

\author{Ziyue Wang\textsuperscript{*,1}, Chi Chen\textsuperscript{*,1}, Yiqi Zhu\textsuperscript{1}, Fuwen Luo\textsuperscript{1}, \\
{\bf Peng Li\textsuperscript{\Letter 2,4}, Ming Yan\textsuperscript{3}, Ji Zhang\textsuperscript{3}, Fei Huang\textsuperscript{\Letter 3}, Maosong Sun\textsuperscript{1}, Yang Liu\textsuperscript{1,2,4,5}} \\
  \textsuperscript{1}1. Dept. of Comp. Sci. \& Tech., Institute for AI, Tsinghua University, Beijing, China \\
  \textsuperscript{2}Institute for AI Industry Research (AIR), Tsinghua University, Beijing, China \\
  \textsuperscript{3}Institute of Intelligent Computing, Alibaba Group\\
  \textsuperscript{4}Shanghai Artificial Intelligence Laboratory, Shanghai, China\\
  \textsuperscript{5}Jiangsu Collaborative Innovation Center for Language Competence, Jiangsu, China
  }

\DefineFNsymbols*{1}{*}
\setfnsymbol{1}

\begin{document}
\maketitle

\renewcommand{\thefootnote}{\fnsymbol{footnote}} 
    \footnotetext[1]{These authors contribute equally.}
\renewcommand{\thefootnote}{\arabic{footnote}}

\DefineFNsymbols*{1}{\Letter}
\setfnsymbol{1}

\renewcommand{\thefootnote}{\fnsymbol{footnote}} 
    \footnotetext[1]{Corresponding authors: Peng Li and Fei Huang.}
\renewcommand{\thefootnote}{\arabic{footnote}}

\begin{abstract}
With the bloom of Large Language Models (LLMs), Multimodal Large Language Models (MLLMs) that incorporate LLMs with pre-trained vision models have recently demonstrated impressive performance across diverse vision-language tasks. However, they fall short to comprehend context involving multiple images. A primary reason for this shortcoming is that the visual features for each images are encoded individually by frozen encoders before feeding into the LLM backbone, lacking awareness of other images and the multimodal instructions.
We term this issue as prior-LLM modality isolation and propose a two phase paradigm, browse-and-concentrate\footnote{Code is released at \url{https://github.com/THUNLP-MT/Brote}}, to enable in-depth multimodal context fusion prior to feeding the features into LLMs.
This paradigm initially ``browses'' through the inputs for essential insights, and then revisits the inputs to ``concentrate'' on crucial details, guided by these insights, to achieve a more comprehensive understanding of the multimodal inputs. 
Additionally, we develop training strategies specifically to enhance the understanding of multi-image inputs.
Our method markedly boosts the performance on 7 multi-image scenarios, contributing to increments on average accuracy by 2.13\% and 7.60\% against strong MLLMs baselines with 3B and 11B LLMs, respectively.
\end{abstract}

\section{Introduction}

Multimodal Large Language Models (MLLMs) have recently garnered attention for their surging popularity and impressive performance across diverse Vision-Language (VL) tasks~\cite{team2023gemini, OpenAI2023Gpt4v, qi2023gemini}. Among these MLLMs, the paradigm that extending Large Language Models (LLMs) with pre-trained vision encoders has shown remarkable abilities in visual reasoning and visual instruction-following~\cite{wu2023multimodal, yin2023survey}. These models also draw attention for their feasibility and flexibility in adapting to varied scenarios and demands~\cite{liu2023medical, zhu2023minigpt}.

\begin{figure}[t]
    \centering
    \includegraphics[width=.5\textwidth]{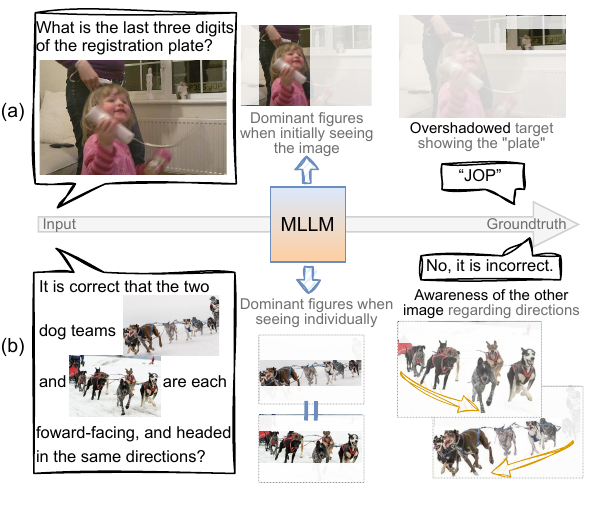}
    \vspace{-2em}
    \caption{Examples of the modality isolation issue. (a) illustrates image-text isolation, where the child figure dominates the image while the ``registration plate'', which should have been focused on, is overshadowed. (b) illustrates inter-image isolation, where the two images lack information regarding ``directions'' of each other. Both situations undergo absence of awareness regarding the global multimodal context.}
    \vspace{-1em}
    \label{fig:1}
\end{figure}

\label{sec:intro}

Despite its impressive abilities, this paradigm faces challenges that obscure a deeper understanding of multi-image and interleaved inputs~\cite{instructblip, luo2023cheap, zhao2023mmicl, li2023fine}. The approach of simply gluing up pre-trained vision and language models via intermediate components~\cite{li2023blip, liu2023visual} potentially neglects essential cross-modality and inter-image interactions, leading to the LLM being presented with isolate visual and textual features without recognition of interleaved multimodal inputs.  We refer to this problem as \emph{prior-LLM modality isolation}, which further divides two issues \emph{image-text isolation} and \emph{inter-image isolation}. These challenges have received considerable attention but remain unresolved.

Firstly, \emph{image-text isolation} happens when frozen vision encoders produce generic visual features, overlooking crucial target-specific information. For instance, in Figure~\ref{fig:1} (a), the emphasis should be on the ``registration plate''. This plate, occupying only a minor area of the image, is prone to being overshadowed by predominant elements due to inadequate image-text interaction. To tackle this problem, \citet{instructblip} and \citet{luo2023cheap} integrate textual instructions into visual feature extraction to enhance the responsiveness of these features to the given instructions. Moreover, some researchers propose to alter the internal structure of LLMs to bridge the gap between visual and linguistic spaces~\cite{wang2023cogvlm}. While these methods are effective in single-image scenarios, they do not address the concurrent fusion of multiple images.

Secondly, \emph{inter-image isolation} arises from encoding images separately, disrupting semantic links among images and conveying misinformation of the multi-image context. This issue is particularly prevalent in scenarios involving interleaved and multiple images. As illustrated in Figure~\ref{fig:1} (b), the moving direction regarding the other image should be considered. However, such relational information remains isolated and fails to transmit across images. Consequently, the lack of awareness regarding relevant content from other images can lead to the exclusion of essential visual information. To handle this issue, recent studies have developed context schemes that aim to improve image-text correlations and the connections between multiple images~\cite{zhao2023mmicl,li2023otter}. Nevertheless, the prior-LLM fusion of multimodal context are overlooked.  

To mitigate the two outlined issues, we utilize a cognitive strategy that mirrors the process through which humans typically understand new content: by first grasping the main ideas during an initial browsing and then revisiting the material to deepen their understanding with the browsing insights~\cite{garner1987metacognition}. Inspired by this approach, we propose a novel paradigm named \textbf{Bro}wse-and-Concentra\textbf{te} (\textbf{Brote}). This paradigm begins with a browsing phase to generate a condition context vector, serving as a collection of browsing insights, encapsulating the main intent and visual information derived from images. Subsequently, a concentrating phase is employed to comprehend multimodal inputs, guided by the condition context vector. Furthermore, to enhance the effectiveness of the browsing insights, we have developed training strategies that prompt the model to implicitly leverage these insights for more precise extraction of image features, allowing for the possibility of bypassing explicit browsing in some scenarios. Our contributions can be summarized as follows:
\vspace{-0.5em}
\begin{itemize}
    \setlength{\itemsep}{0pt}
    \setlength{\parsep}{0pt}
    \setlength{\parskip}{0pt}
    \item We address the challenge of prior-LLM modality isolation by proposing the browse-and-concentrate paradigm, alongside training strategies to encourage the model to leverage and explore the browsing insights.
    \item We explore two method to implement our paradigm, demonstrating that Brote not only learns to concentrate on interleaved inputs via explicit context vectors, but also integrates this ability directly into the model implicitly. 
    \item We conduct comprehensive evaluations on 7 multi-image scenarios and exhibits notable advancements, improving the average accuracy by 2.13\% and 7.60\% against baselines with 3B and 11B LLMs, respectively. 
\end{itemize}
\vspace{-0.5em}

\begin{figure*}[!th] 
\begin{center}
\includegraphics[width=.98\textwidth]{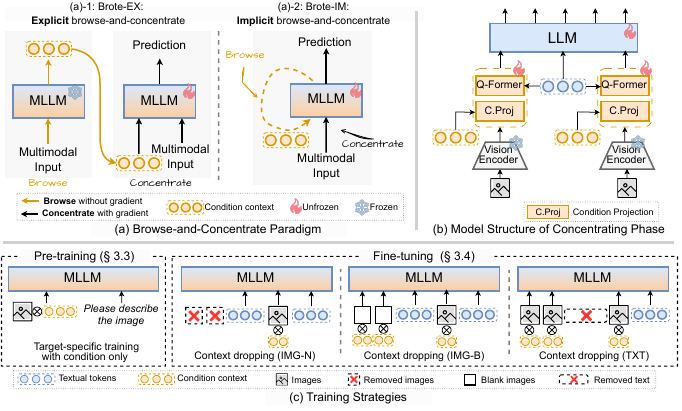}
\vspace{-0.5em}
\caption{The illustration of browse-and-concentrate paradigm (a), model architecture of the concentrating phase (b), and our proposed training strategies (c). (a) shows the pipelines of Brote models, (a)-1 for Brote-EX and (a)-2 for Brote-IM. (c) depicts the strategies described in \S\ref{sec:pre_strategy} and \S\ref{sec:fine_strategy}.} \label{fig:model}
\vspace{-1em}
\end{center}
\end{figure*}

\section{Related Work}

\subsection{Empowering LLMs with Visual Abilities via Pre-trained Vision Models}
With the surging of LLMs, MLLMs that empower LLMs with visual abilities have also witnessed a rapid growth. Following the initial effort~\cite{tsimpoukelli2021multimodal} to convert visual features into readable embeddings for LLMs, researchers have proposed to bridge vision and language modalities via diverse visual prompt generators (VPG), such as Resampler~\cite{alayrac2022flamingo}, Q-Former~\cite{li2023blip, instructblip}, and linear projections~\cite{liu2023visual, huang2023language}. They utilize image features from frozen vision models~\cite{dosovitskiy2020image, radford2021learning}, and subsequently integrate these features into pre-trained LLMs. These MLLMs inherit cognitive and perceptual abilities from vision models and the emergent ability from LLMs, exhibiting impressive performance without intensive training. 
However, they bear the modality isolation issue that obscures a deeper understanding multimodal context.

\subsection{Enhancing Visual Features with Textual Instructions}
Recent studies have concentrated on augmenting the capability for MLLMs to follow visual instructions ~\cite{liu2023visual, instructblip, luo2023cheap, wang2023cogvlm, ye2023mplug}. Some researchers have focused on the fine-tuning LLMs to better response to visual instructions~\cite{ye2023mplug, liu2023improved}, employing techniques such as LoRA~\cite{hu2021lora}. While other studies target the issue of image-text isolation by manipulating with the visual features. For instance, \citet{instructblip} enhance Q-Former with textual instructions to obtain instruction-aware visual features. \citet{luo2023cheap} integrate learnable instruction features directly into the vision encoders. Despite these innovations, they primarily incorporate instructions into the vision modules pay less attention to the complexity of multi-image senarios. 

\subsection{MLLMs Enhanced for Comprehending Multiple Images}
The ability to comprehend multiple images simultaneously draws considerable attention~\cite{alayrac2022flamingo, zhao2023mmicl, li2023otter, shukor2023beyond}. Multi-image scenarios can be categorized into interleaved image-text formats and multimodal ICL settings. To improve ICL preformance, \citet{shukor2023beyond} analyse prompt-based approaches, introducing three templates for multimodal ICL. Meanwhile, some researchers work on methods requiring model-tuning~\cite{alayrac2022flamingo, shukor2023beyond, sun2023generative}. Additionally, some scholars broaden the exploration of multi-image scenarios to include both ICL and interleaved inputs. \citet{li2023fine} insert middle-layer LLM outputs into the VPG as additional guidance for spotting the differences between images. \citet{zhao2023mmicl} and \citet{li2023otter} construct datasets targeting the multi-image issue, and propose context schemes to improve the understanding of interleaved inputs. 
Despite these advancements, the prior-LLM multimodal context fusion is not sufficiently explored.

\section{Method}
\subsection{Overview} \label{sec:overview}
To stimulate prior-LLM multimodal context fusion and improve the awareness of multimodal context of the LLM, we propose a paradigm, \textbf{Bro}wse-and-concentra\textbf{te} (Brote). It progressively comprehends images via two phases, \textbf{browsing} and \textbf{concentrating}. As illustrated in Figure~\ref{fig:model} (a), in the browsing phase, the MLLM browses the entire input and generates a condition context as the browsing result, denoted as $\mathcal{C}$ in the rest of this paper. Then, in the concentrating phase, the model comprehends multimodal inputs under the guidance of $\mathcal{C}$. We refer to the model of browsing phase as $\mathcal{M}_{B}$ and the model of concentrating phase as $\mathcal{M}_{C}$.

Our proposed Brote can be further divided into two modes, \emph{explicit} and \emph{implicit}, regarding the distinct approaches of incorporating $\mathcal{C}$. The explicit browse-and-concentrate (Figure~\ref{fig:model} (a)-1), denoted as \textit{Brote-EX}, operates with separated parameters ($\mathcal{M}_{B} \neq \mathcal{M}_{C}$). This explicit mode first generates $\mathcal{C}$ using $\mathcal{M}_{B}$, followed by $\mathcal{M}_{C}$ to infer the final outcomes. In contrast, for the implicit browse-and-concentrate (Figure~\ref{fig:model} (a)-2), denoted as \textit{Brote-IM}, employs shared parameters for both phases ($\mathcal{M}_{B} = \mathcal{M}_{C}$), permitting $\mathcal{M}_{C}$ to directly predict the answer without the need to explicitly produce intermediate vectors from the other model. 
Along with the proposed paradigm, we devise training strategies for the explicit browse-and-concentrate mode. This strategies encourage the model to leverage and explore the generated condition context vectors. 
The explicit mode serves as a precursor to the implicit mode, preparing the model with fundamental and essential ability to understand $\mathcal{C}$.

We will elaborately describe the workflow of Brote in \S\ref{sec:workflow}, followed by the proposed strategies for pre-training (\S\ref{sec:pre_strategy}) and fine-tuning (\S\ref{sec:fine_strategy}). 

\subsection{Browse-and-Concentrate Paradigm}\label{sec:workflow}
We represent the interleaved multimodal input as $\bm{x}$, defined as $\bm{x} = [x^{m}_0, x^{m}_1, \dots, x^{m}_n,\dots, x^{m}_{N-1}]$ for $N$ tokens, with $n=0,1,\dots,{N-1}$. Each token is associated with modality $m$, where $m \in \{\mathrm{image}, \mathrm{text}\}$. 
Images are individually encoded by vision encoder $g_{\phi_{v}}(\cdot)$ with parameters $\phi_{v}$, which provides image features $\bm{v}=g_{\phi_{v}}(x^m_{n})$, for $m=\mathrm{image}$. Referring to $\mathrm{Emb}(\cdot)$ as the embedding mapping, $\bm{v}$ is subsequently integrated with textual instructions $\bm{h}^{\mathrm{text}}=\mathrm{Emb}(x_{n}^{m})$, for $m=\mathrm{text}$, via a Q-Former $f_{\phi_{Q}}(\cdot,\cdot)$ parameterized by $\phi_{Q}$,
\begin{align}
    \bm{h} & = [h^{m}_0, h^{m}_1, \dots, h^{m}_n,\dots, h^{m}_{N-1}]  \\
    \bm{h}^{m}_{n} & = 
    \begin{cases}
         \mathrm{Emb}(x_{n}^{m}) & \text{ if } m=\mathrm{text} \\
         f_{\phi_{Q}}(\bm{v}, [\bm{Q};\bm{h}^{\mathrm{text}}]) & \text{ if } m=\mathrm{image} \\
    \end{cases}
\end{align}
where $\bm{h}$ denotes the multimodal embeddings, and $\bm{Q}$ is the learnable query tokens in Q-Former. 

The LLM component of $\mathcal{M}_{B}$ is denoted by $f_{\phi_L}(\cdot)$ with parameters $\phi_{L}$. The browsing phase produces $\mathcal{C}$ by extracting the last hidden states of the LLM $f_{\phi_L}(\cdot)$, denoting as follows:
\begin{align}
    \mathcal{C} =f_{\phi_L}^{(l)}(\bm{h}). 
\end{align}
where $l$ represents the last layer of $f_{\phi_L}(\cdot)$.

In the concentrating phase, the images undergo alterations conditioned on $\mathcal{C}$. We add $\mathcal{C}$ to query tokens $Q$, and obtain the altered visual token embeddings $\tilde{\bm{h}}^{\mathrm{image}}$ as,
\begin{align}
    \tilde{\bm{h}}^{\mathrm{image}} &= f_{\phi_{Q}}(\bm{v}, [\bm{Q}+\mathrm{Linear}(\mathcal{C});\bm{h}^{\mathrm{text}}]),
\end{align}
where $\mathrm{Linear}(\cdot)$ denotes the linear projection, and $[\cdot;\cdot]$ denotes concatenation. In this phase, Q-Former accepts an extra input $\mathcal{C}$ compare to the browsing phase. Finally, the prediction $\bm{y}$ with $T$ tokens is formulated as follows:
\begin{align}\label{eq:m_c}
    \bm{y} &= \mathcal{M}_{C}(\bm{x}, \mathcal{C}) \notag \\
    &= \mathop{\mathrm{argmax}}_{\bm{y}}p(y|\bm{x}, \mathcal{C}; f_{\phi_L'}, f_{\phi_Q'}, g_{\phi_v'}) \\\vspace{-0.5em}
    &= \mathop{\mathrm{argmax}}_{\bm{y}} \prod_{t=1}^{T} p(y_{t}|y_{<t}, \bm{x}, \mathcal{C}; f_{\phi_L'}, f_{\phi_Q'}, g_{\phi_v'}), \notag \vspace{-0.5em}
\end{align}
where $y_{t}$ is the $t$-th token of the prediction, $y_{<t}=y_1,\cdots,y_{t-1}$, $f_{\phi_L'}$, $f_{\phi_Q'}$, $g_{\phi_v'}$ are the LLM, Q-Former, and vision encoder components of $\mathcal{M}_C$ parameterized by $\phi_L'$, $\phi_Q'$, and $\phi_v'$, respectively.

\subsection{Context-Enhanced Pre-Training} \label{sec:pre_strategy}

\begin{figure}
    \centering
    \includegraphics[width=.47\textwidth]{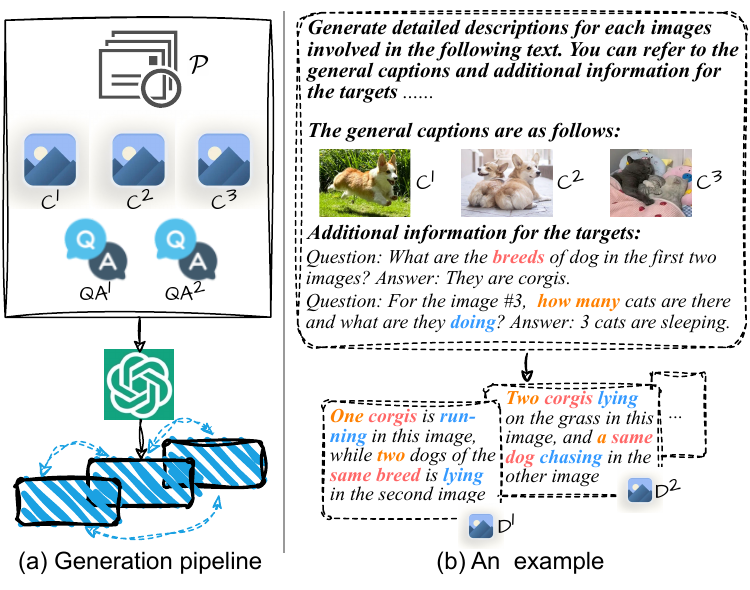}
    \vspace{-0.5em}
    \caption{A illustration of data construction process (a) described in \S\ref{sec:data}, with a detailed example (b). The generated descriptions should be aware of both the targets (``\textit{breeds}'', ``\textit{how many}'', ``\textit{doing}'') and another images. }
    \label{fig:data_con}
\end{figure}

\paragraph{Condition context-enhanced pre-training.} \label{sec:pretrain}
The pre-training stage aims at adapting the model to utilize $\mathcal{C}$ and enhancing visual feature extraction with its conveyed multimodal context. To this end, we propose a training task that challenges the model to generate task-specific descriptions without direct exposure to the question. Initially, we obtain $\mathcal{C}$ by feeding the intact inputs into $\mathcal{M}_{B}$. Then, in the concentrating phase, $\mathcal{M}_{C}$ is required to generate image descriptions specialised for the questions that are not explicitly presented but instead implicitly encoded within $\mathcal{C}$. As depicted in Figure~\ref{fig:model} (c) ``Pre-training'', the model is presented with only the text ``\textit{Please describe the image}'' alongside altered visual tokens $\tilde{\bm{h}}^{\mathrm{image}}$. This strategy urges the model to explore $\mathcal{C}$ for target information. Additionally, we combine the task-specific training targets together with the general ones, enabling the model to discern between inputs with and without $\mathcal{C}$. The objective for pre-training is as follows:
\begin{align} \label{eq:loss_m_c}
    \!\mathcal{L}_{\mathcal{M}_{C}}\!=\!-\!\sum_{t=1}^{T} \hat{y}_t \log p(y_{t} | \bm{x},\mathcal{C}; f_{\phi_L'}, f_{\phi_Q'}, g_{\phi_v'}),\hspace{-0.5em} \vspace{-0.5em}
\end{align}
where $\hat{y}_{t}$ is the $t$-th groundtruth token.

\paragraph{Data construction.} \label{sec:data}
In alignment with the task-specific training strategy, we design a data generation method to secure task-specific supervisions as mentioned above. Inspired by PromptCap~\cite{hu2023promptcap}, we leverage LLMs to craft target-aware image descriptions. Our approach is extended from the producing of individual image descriptions to addressing multiple interleaved inputs, enabling a more profound understanding of multi-image and interleaved context. We obtain the image-target related descriptions as demonstrated in Figure~\ref{fig:data_con}. The LLM receives a triplet $(\mathcal{P}, C^{K}, QA^{J})$, comprising task instruction prompt $\mathcal{P}$, general image descriptions $C^{K}$ and question-answer pairs $QA^{J}$, where $K$ and $J$ represent the counts of images and targeted question-answer pairs respectively. Also noticing that $K$ is not necessarily equal to $J$. For each $k$-th image ($k=1,2,\dots,K$), the LLM is required to generate image description $D^k$ that satisfies the target clarified in $\mathcal{P}$. Accordingly, $D^k$ contains specific messages for questions in $QA^{J}$ and information about the other related images $o$, $k \neq o$. 

We construct a total of 56k data for the pre-training stage, and manually assess the quality of the generated captions by randomly sampling 230 generated captions. We detect that 36 (out of 230) captions contain hallucination or minor incorrect information, while the rest 84\% are of good quality, containing desired and correct question-aware information. Please refer to Appendix~\ref{appendix:stage1data} for details of the generated data.

\subsection{Condition-Aware Task Fine-Tuning} \label{sec:fine_strategy}
To encourage further exploration of information from $\mathcal{C}$ for VL tasks, we propose a new training strategy named \textit{context-dropping training}. The strategy intentionally omits particular inputs yet requiring the model to infer for answers solely with the assistant of $\mathcal{C}$. It motivates the model to compensate for the missing  information from the provided condition context $\mathcal{C}$. We propose different dropping strategies as illustrated in Figure~\ref{fig:model} (b): 
\vspace{-0.5em}
\begin{itemize}
    \setlength{\itemsep}{0pt}
    \setlength{\parsep}{0pt}
    \setlength{\parskip}{0pt}
    \item Drop images: This involves two approaches, removing certain images (Figure~\ref{fig:model} (b), ``Context Dropping (IMG-N)''), and replacing original images by blank placeholders (Figure~\ref{fig:model} (b), ``Context Dropping (IMG-B)'').
    \item Drop text: We remove the text before the last image as shown in Figure~\ref{fig:model} (b), ``Context Dropping (TXT)''.
    \item Drop ALL: A combination of the above settings denoted as ``ALL'', applied with the same probabilities.
    \vspace{-0.5em}
\end{itemize}
To ensure integration with $\mathcal{C}$, we preserve the last image across all dropping strategies. Notice that the ``drop images'' approaches are not applicable to inputs with only one image. These strategies compel the model to infer indispensable information from $\mathcal{C}$ that should have been given in the input. 

As mentioned in \S\ref{sec:overview}, we investigate two modes for incorporating $\mathcal{C}$, Brote-EX and Brote-IM. For Brote-EX, we apply context-dropping strategies to the concentrating phase with $\mathcal{C}$ provided by frozen model $\mathcal{M}_{B}$. The training objective for explicit mode is $\mathcal{L}_{\mathcal{M}_{C}}$ as described in Equation~\ref{eq:loss_m_c}. While for Brote-IM, parameters of $\mathcal{M}_{B}$ are shared with $\mathcal{M}_{C}$. When optimizing the shared parameters, we also take into account the loss for $\mathcal{M}_{B}$ as follows:
\begin{align}
    \mathcal{L}_{\mathcal{M}_{B}} = -\sum_{t=1}^{T} \hat{y}_t \log p(y_{t} | \bm{x}; f_{\phi_L}, f_{\phi_{Q}}, g_{\phi_{v}}). 
\end{align}
For the training of Brote-IM, we sum up the two losses, for $\mathcal{M}_{B}$ and $\mathcal{M}_{C}$ respectively, as $\mathcal{L}_{\mathcal{M}_{B}}+\mathcal{L}_{\mathcal{M}_{C}}$, denoted by \textit{dual-loss}. Details of the training process are documented in Appendix~\ref{appendix:training}.

\begin{table*}[!ht]
\centering\footnotesize
\begin{tabular}{l@{\hspace{0.25cm}}l@{\hspace{0.1cm}}c|c@{\hspace{0.1cm}}cc@{\hspace{0.2cm}}c@{\hspace{0.2cm}}c@{\hspace{0.2cm}}c@{\hspace{0.2cm}}c|c}
\toprule 
& \multirow{2}{*}[-2ex]{Model} & \multirow{2}{*}[-2ex]  {\begin{tabular}[c]{@{}c@{}}\#Param\\LLM\end{tabular}} & \multicolumn{2}{c}{In-context Learning} &  \multicolumn{5}{c|}{Multi-image / Video Tasks} & \multirow{2}{*}[-2ex]{AVG} \\\cmidrule(lr){4-5}\cmidrule(lr){6-10}
 & & & VQAv2 & A-OKVQA & NLVR2 & DEMON & SEED & \begin{tabular}[c]{@{}c@{}}MSVD\\QA\end{tabular} & \begin{tabular}[c]{@{}c@{}}MSRVTT\\QA\end{tabular} & \\ \midrule 
\multirow{8}{*}{\rotatebox{90}{\scriptsize{\#LLM $\leq$ 10B}}} & \textcolor{gray}{KOSMOS-1} & \textcolor{gray}{1.3B} & \textcolor{gray}{51.8}\hphantom{0$^*$} & \textcolor{gray}{-} & \textcolor{gray}{-} & \textcolor{gray}{-} & \textcolor{gray}{-} & \textcolor{gray}{-} & \textcolor{gray}{-} & \textcolor{gray}{-} \\
& InstructBLIP-XL &   3B    & 31.76$^*$ & 39.13$^*$ & 52.59$^*$ & 32.59$^*$ & 52.7\hphantom{0$^*$} & 43.40$^*$ & 12.12$^*$ & 37.77  \\  
& MMICL-XL$^{\diamondsuit}$  &   3B  & 69.16\hphantom{$^*$} &  53.43$^*$ & \underline{71.48}$^*$ & \textbf{38.14}$^*$ & 54.69$^*$ & \underline{53.68}\hphantom{$^*$} & 42.36$^*$ & 54.71 \\ 
& \textcolor{gray}{Otter}      &   \textcolor{gray}{7B}   & \textcolor{gray}{45.39$^*$} & \textcolor{gray}{38.42$^*$} & \textcolor{gray}{49.54$^*$} & \textcolor{gray}{24.51}\hphantom{$^*$} & \textcolor{gray}{39.7}\hphantom{0$^*$} & \textcolor{gray}{25.87$^*$} & \textcolor{gray}{\hphantom{0}9.78$^*$} & \textcolor{gray}{33.32} \\
& \textcolor{gray}{VPG-C-LLaMA2}  &  \textcolor{gray}{7B} & \textcolor{gray}{-} & \textcolor{gray}{34.29$^*$} & \textcolor{gray}{53.82$^*$} & \textcolor{gray}{37.22}\hphantom{$^*$} & \textcolor{gray}{-} & \hphantom{0}\textcolor{gray}{6.03$^*$} & \textcolor{gray}{-} & \textcolor{gray}{-} \\ 
& \textcolor{gray}{Flamingo-9B} &  \textcolor{gray}{7B} & \color{gray} 56.3\hphantom{0$^*$} & \textcolor{gray}{-} & \textcolor{gray}{-} & \textcolor{gray}{-} & \textcolor{gray}{-} & \textcolor{gray}{30.2\hphantom{0$^*$}} & \textcolor{gray}{13.7\hphantom{0$^*$}} & \textcolor{gray}{-} \\ 
\noalign{\vskip 1pt} \cdashline{2-11} \noalign{\vskip 2pt}
& Brote-EX-XL (ours)       &   3B  & \textbf{69.97}\hphantom{$^*$}  & \underline{56.00}\hphantom{$^*$}  & 71.41\hphantom{$^*$} & 37.33\hphantom{$^*$} & \underline{57.51}\hphantom{$^*$} & 53.02\hphantom{$^*$} & \underline{43.14}\hphantom{$^*$} & \underline{55.48} \\
& Brote-IM-XL (ours)  &   3B  & \underline{68.94}\hphantom{$^*$} &  \textbf{56.43}\hphantom{$^*$}  & \textbf{76.02}\hphantom{$^*$} & \underline{37.34}\hphantom{$^*$} & \textbf{57.86}\hphantom{$^*$} & \textbf{56.06}\hphantom{$^*$} & \textbf{45.08}\hphantom{$^*$} & \textbf{56.84}   \\
\midrule 
\multirow{6}{*}{\rotatebox{90}{\scriptsize{\#LLM $>$ 10B}}} & InstructBLIP-XXL &   11B    & 48.21$^*$ & 45.92$^*$ & 64.54$^*$ & 33.00$^*$ & 50.81$^*$ & 44.30$^*$ & 15.49$^*$ & 43.18  \\  
& MMICL-XXL$^{\diamondsuit}$   &    11B    & 70.56\hphantom{$^*$}  & 54.85$^*$ & 56.16$^*$ & 36.30$^*$ & 56.66$^*$ & 52.19\hphantom{$^*$} & 39.46$^*$ & 52.18 \\
& \textcolor{gray}{EMU-2}   &    \textcolor{gray}{33B}    & \textcolor{gray}{67.0}\hphantom{0$^*$} & \textcolor{gray}{-} & \textcolor{gray}{-} & \textcolor{gray}{-} & \color{gray} \textbf{62.8\hphantom{0$^*$}} & \textcolor{gray}{49.0\hphantom{0$^*$}} & \textcolor{gray}{31.4\hphantom{0$^*$}} & \textcolor{gray}{-} \\
& \textcolor{gray}{Flamingo-80B} &  \textcolor{gray}{70B} & \textcolor{gray}{63.1\hphantom{0$^*$}} & \textcolor{gray}{-} & \textcolor{gray}{-} & \textcolor{gray}{-} & \textcolor{gray}{-} & \textcolor{gray}{35.6\hphantom{0$^*$}} & \textcolor{gray}{17.4\hphantom{0$^*$}} & \textcolor{gray}{-} \\ 
\noalign{\vskip 1pt} \cdashline{2-11} \noalign{\vskip 2pt}
& Brote-EX-XXL (ours)    &    11B  & \underline{70.86}\hphantom{$^*$} & \underline{59.94}\hphantom{$^*$}  & \underline{70.42}\hphantom{$^*$} & \underline{38.70}\hphantom{$^*$} & 59.31\hphantom{$^*$} & \underline{54.52}\hphantom{$^*$} & \underline{45.24}\hphantom{$^*$} & \underline{57.00} \\ 
& Brote-IM-XXL (ours)  &  11B  & \textbf{71.71}\hphantom{$^*$} & \textbf{60.31}\hphantom{$^*$} & \textbf{80.71}\hphantom{$^*$} & \textbf{38.94}\hphantom{$^*$} & \underline{61.64}\hphantom{$^*$} & \textbf{57.29}\hphantom{$^*$} & \textbf{45.94}\hphantom{$^*$} & \textbf{59.78} \\ 
\bottomrule
\end{tabular}
\vspace{-0.1cm}
\caption{Results for multi-image settings. The best results for models larger/smaller than 10B are separately \textbf{bolded} and the seconds are \underline{underlined}. $^{\diamondsuit}$: the InstructBLIP version. We evaluate results which are not officially announced using public checkpoints and mark them by *. SEED refers to SEED-Bench that contains both images and videos.} 
\label{tab:main_multi}
\end{table*}

\begin{table*}[!ht]
\centering\footnotesize
\begin{tabular}{l@{\hspace{0.25cm}}l@{\hspace{0.1cm}}c|c@{\hspace{0.2cm}}c@{\hspace{0.2cm}}c@{\hspace{0.2cm}}c@{\hspace{0.2cm}}c@{\hspace{0.2cm}}c|c}
\toprule 
& Model & \begin{tabular}[c]{@{}c@{}}\#Param\\LLM\end{tabular}& VQAv2 & A-OKVQA & \begin{tabular}[c]{@{}c@{}}ScienceQA\\-IMG\end{tabular} & \begin{tabular}[c]{@{}c@{}}MME\\Perception\end{tabular} & \begin{tabular}[c]{@{}c@{}}MME\\Cognition\end{tabular}  & MMBench & AVG\\ \midrule 
\multirow{6}{*}{\rotatebox{90}{\scriptsize{\#LLM $\leq$ 10B}}} & InstructBLIP-XL & 3B & 36.77\hphantom{$^*$} & \textbf{54.57}\hphantom{$^*$} & 70.40\hphantom{$^*$} & 1093.70$^*$ & 281.43$^*$ &  69.68$^*$  & 68.52 \\  
 & MMICL-XL      &   3B & 69.13\hphantom{$^*$} &  52.12$^*$ & \textbf{72.58}$^*$ & 1184.54$^*$ & 277.86$^*$  & 73.11$^*$  & 75.81 \\ 
 &  \color{gray} LLaVA$^\dag$ & \color{gray} 7B & \color{gray} - &\color{gray} - & \color{gray}- & \color{gray}457.82 & \color{gray}214.64\hphantom{$^*$} & \color{gray}36.2\hphantom{0$^*$} & \textcolor{gray} - \\
 & \color{gray}Otter$^\dag$      &   \color{gray}7B   & \color{gray} 57.89$^*$ &\color{gray}41.92$^*$ & \color{gray}63.10\hphantom{$^*$} & \color{gray}\textbf{1292.26}\hphantom{$^*$} & \color{gray}\textbf{306.43}\hphantom{$^*$} & \color{gray}48.3\hphantom{0$^*$} & 69.51 \\
\noalign{\vskip 1pt} \cdashline{2-10} \noalign{\vskip 2pt}
 & Brote-EX-XL (ours)   &  3B & \underline{69.90}\hphantom{$^*$} & 52.93\hphantom{$^*$}  & \underline{71.15}\hphantom{$^*$}  & \underline{1203.87}\hphantom{$^*$} & \underline{301.79}\hphantom{$^*$} & \underline{73.27}\hphantom{$^*$} & \textbf{77.18} \\ 
 & Brote-IM-XL (ours)   &  3B & \textbf{70.24}\hphantom{$^*$} & \underline{53.40}\hphantom{$^*$} & \textbf{72.58}\hphantom{$^*$}  & 1181.95\hphantom{$^*$} & 266.79\hphantom{$^*$} & \textbf{74.29}\hphantom{$^*$} & \underline{75.90} \\ \midrule 
\multirow{5}{*}{\rotatebox{90}{\scriptsize{\#LLM $>$ 10B}}}  & InstructBLIP-XXL &   11B    & 63.69\hphantom{$^*$} & \underline{57.10}\hphantom{$^*$} & 70.60\hphantom{$^*$} & 1212.82$^*$ & 291.79$^*$ & 70.34$^*$ & 75.99  \\  
& MMICL-XXL & 11B  & 70.30\hphantom{$^*$}  & 51.35$^*$ & 74.92$^*$ & \underline{1313.8}8$^*$ & \underline{311.79}$^*$ & 76.58$^*$  & 80.41 \\
& MMICL-XXL (BLIP-2)$^\dag$  &  11B  & 69.99\hphantom{$^*$} & - & - & \textbf{1381.74}\hphantom{$^*$} & \textbf{428.93}\hphantom{$^*$} & 65.24\hphantom{$^*$}  & \textcolor{gray} - \\
\noalign{\vskip 1pt} \cdashline{2-10} \noalign{\vskip 2pt}
& Brote-EX-XXL (ours)   &  11B  & \underline{71.58}\hphantom{$^*$} & 56.47\hphantom{$^*$} & \underline{77.69}\hphantom{$^*$} & 1279.73\hphantom{$^*$} & 310.01\hphantom{$^*$} & \underline{76.67}\hphantom{$^*$} & \underline{81.31} \\ 
& Brote-IM-XXL (ours)   &  11B  & \textbf{73.02}\hphantom{$^*$} & \textbf{57.83}\hphantom{$^*$} & \textbf{78.38}\hphantom{$^*$} & 1284.13\hphantom{$^*$} & 300.00\hphantom{$^*$} & \textbf{77.34}\hphantom{$^*$} & \textbf{81.66} \\ 
\bottomrule
\end{tabular}
\vspace{-0.1cm}
\caption{Zero-shot results for single-image settings. The best results for models larger/smaller than 10B are separately \textbf{bolded} and the seconds are \underline{underlined}. 
$^\dag$: results of these models are taken from \citet{zhao2023mmicl}. We evaluate results which are not officially announced using public checkpoints and mark them by *. For ``AVG'', we first average the MME scores over its subtasks, then calculate the average scores of all benchmarks in this table. We include closely related baselines in this table, and refer readers to Appendix~\ref{appendix:full_rst} for detailed results of other models.} 
\vspace{-1em}
\label{tab:main_single}
\end{table*}

\section{Experiments}
\subsection{Implementation}
We implement our method upon InstructBLIP~\cite{instructblip} with FlanT5~\cite{chung2022scaling} as the language backbone. We pre-train our model on the 56k generated data as described in \S\ref{sec:pre_strategy}, and then extract about 490k data from the MIC dataset~\cite{zhao2023mmicl} for model fine-tuning. The fine-tuning data is sampled according to the data-balanced sampling algorithm suggested by \citet{instructblip}. Please refer to Appendix~\ref{appendix:trainingdata} for details of the training data and Appendix~\ref{appendix:training} for more information of the training process. 

\subsection{Evaluation Settings}
\paragraph{Baselines.} 
We primarily employ models designed for accepting multiple images or interleaved image-text inputs as baselines, such as MMICL~\cite{zhao2023mmicl}, Otter~\cite{li2023otter} and VPG-C~\cite{li2023fine}. Additionally, MLLMs that are used to develop these baselines are also considered, such as BLIP-2~\cite{li2023blip} and InstructBLIP~\cite{instructblip}. Please refer to Appendix~\ref{appendix:baseline} for detailed information of the employed baselines. For models whose results are not officially reported, we utilize the publicly available checkpoints for evaluation.\footnote{We use the public checkpoints to obtain the missing results for MMICL (\url{https://huggingface.co/BleachNick}), InstructBLIP (\url{https://huggingface.co/Salesforce}), and Otter (\url{https://huggingface.co/luodian}), together with official scripts and required environments.}  

\paragraph{Benchmarks and Metrics.} 
We investigate diverse VL benchmarks and focus on multi-image tasks, including visual reasoning (NLVR2~\cite{suhr2019corpus}), few-shot ICL for image QA (VQAv2~\cite{goyal2017making} and A-OKVQA~\cite{schwenk2022aok}), video QA (MSVD QA~\cite{xu2017video}, MSRVTT QA~\cite{xu2017video}, SEED-Bench~\cite{li2023seedbench}), and multi-image instruction following (DEMON~\cite{li2023fine}).
Note that SEED-Bench comprises of both images and videos. For video benchmarks, following \citet{zhao2023mmicl}, we uniformly extract eight frames from the given video clips for answering the questions. For few-shot ICL, we employ the widely used four-shot setting. Additionally, we conduct experiments on single-image tasks to fairly compare with models that are not designed for multi-image settings. These tasks include zero-shot setting for VQAv2, A-OKVQA and MME~\cite{fu2023mme}. ScienceQA~\cite{saikh2022scienceqa} (SciQA) is designed for Chain-of-Thought (CoT)~\cite{wei2022chain} scenario with accompany hints, and we adopt the zero-shot CoT (ZS-CoT) setting for this dataset. Details of these evaluation benchmarks, including data scale, the type of tasks and evaluation metrics, are listed in Appendix~\ref{appendix:metrics}.

\subsection{Results}

We report results for multi-image settings in Table~\ref{tab:main_multi} and single-image settings in Table~\ref{tab:main_single}. Drawing conclusions from these tables, our method presents significant improvement for multi-image settings, while concurrently improves the performance of 3 single-image tasks.

\begin{table}[!t]
\centering
\resizebox{.48\textwidth}{!}{
\begin{tabular}{@{\hspace{0.cm}}l@{\hspace{0.1cm}}c@{\hspace{0.05cm}}c@{\hspace{0.1cm}}|@{\hspace{0.05cm}}c@{\hspace{0.05cm}}|@{\hspace{0.15cm}}l@{\hspace{0.15cm}}l@{\hspace{0.15cm}}|@{\hspace{0.15cm}}l@{\hspace{0.15cm}}l@{\hspace{0cm}}}
\toprule
& \multirow{2}{*}{Dataset} & \multirow{2}{*}{Settings} & \multirow{2}{*}{MMICL} & \multicolumn{2}{@{\hspace{0.15cm}}c@{\hspace{0.15cm}}|@{\hspace{0.15cm}}}{Brote-EX} & \multicolumn{2}{@{\hspace{0.15cm}}c@{\hspace{0.2cm}}}{Brote-IM} \\ 
&  &  & & Ours & Ours-None & Ours & Ours-None \\ \midrule
\multirow{6}{*}{\rotatebox{90}{\scriptsize{XL}}} & \multirow{2}{*}{\begin{tabular}[c]{@{}c@{}}A-\\OKVQA\end{tabular}}  & 0-shot & 52.12 & 52.93 & 49.40 \scriptsize\textcolor{red}{-3.53} & 53.40 & 51.88 \scriptsize\textcolor{red}{-1.52} \\ 
  &  & 4-shot & 53.43 & 56.00 & 55.10 \scriptsize\textcolor{red}{-0.90} & 56.53 & 56.28 \scriptsize\textcolor{red}{-0.25}   \\ 
\noalign{\vskip 1pt} \cdashline{2-8}\noalign{\vskip 2pt}
& \multirow{2}{*}{SEED}  & Image & 57.99 & 61.90 & 59.79 \scriptsize\textcolor{red}{-2.11}& 61.82 &  61.48 \scriptsize\textcolor{red}{-0.34} \\
&    & Video & 41.94 & 40.50 & 39.06 \scriptsize\textcolor{red}{-1.44} & 42.52 &  42.11 \scriptsize\textcolor{red}{-0.41} \\
\noalign{\vskip 1pt} \cdashline{2-8}\noalign{\vskip 2pt}
& NLVR2   & 0-shot & 71.48 & 71.41 & 69.27 \scriptsize\textcolor{red}{-2.14} & 76.02 & 75.59 \scriptsize\textcolor{red}{-0.43} \\ 
\noalign{\vskip 1pt} \cline{2-8}\noalign{\vskip 2pt}
& \multicolumn{2}{c@{\hspace{0.05cm}}|@{\hspace{0.05cm}}}{Average} & 55.39 & 56.55 & 54.52 \scriptsize\textcolor{red}{-2.03} & 58.06 & 57.47 \scriptsize\textcolor{red}{-0.59} \\ 
\midrule
\multirow{6}{*}{\rotatebox{90}{\scriptsize{XXL}}} & \multirow{2}{*}{\begin{tabular}[c]{@{}c@{}}A-\\OKVQA\end{tabular}}  & 0-shot & 51.35 & 56.47 & 55.32 \scriptsize\textcolor{red}{-1.15} & 57.83 & 57.61 \scriptsize\textcolor{red}{-0.22} \\ 
  &  & 4-shot & 54.85 & 59.94 & 58.70 \scriptsize\textcolor{red}{-1.24} & 60.65 & 60.25 \scriptsize\textcolor{red}{-0.40}   \\ 
\noalign{\vskip 1pt} \cdashline{2-8}\noalign{\vskip 2pt}
& \multirow{2}{*}{SEED}  & Image & 59.17 & 63.70 & 63.26 \scriptsize\textcolor{red}{-0.44} & 65.58 &  64.65 \scriptsize\textcolor{red}{-0.93} \\
&    & Video & 46.90 & 42.27 & 41.88 \scriptsize\textcolor{red}{-0.39}& 46.37 &  46.25 \scriptsize\textcolor{red}{-0.12} \\
\noalign{\vskip 1pt} \cdashline{2-8}\noalign{\vskip 2pt}
& NLVR2   & 0-shot & 53.62 & 70.42 & 68.60 \scriptsize\textcolor{red}{-1.82} & 80.71 & 79.69 \scriptsize\textcolor{red}{-1.02} \\ 
\noalign{\vskip 1pt} \cline{2-8}\noalign{\vskip 2pt}
& \multicolumn{2}{c@{\hspace{0.05cm}}|@{\hspace{0.05cm}}}{Average} & 53.18 & 58.56 & 57.55 \scriptsize\textcolor{red}{-1.01} & 62.23 & 61.69 \scriptsize\textcolor{red}{-0.54} \\
\bottomrule
\end{tabular}
}
\vspace{-0.5em}
\caption{Ablation study of the condition context vectors. ``Ours-None'' indicates the none condition setting (replacing the condition by all-zero vectors when testing).} 
\vspace{-1em}
\label{tab:abla_cond}
\end{table}

Our models exhibit notable advancements over models in Table~\ref{tab:main_multi}, showing profound comprehending ability for multi-image and interleaved inputs. We outperform strong baselines, such as InstructBLIP, MMICL and VPG-C, which include shallow prior-LLM instruction-image fusion. Our method goes beyond merely cross-modality integration between image and text to also include intra-modality fusion among images. Impressively, our models show consistent advantage over benchmarks involving videos and multiple images, and for few-shot ICL of QA tasks as well. For the average scores of models following InstructBLIP paradigm, our models achieve improvements of 2.13\% and 7.60\% for XL and XXL models respectively, over MMICL. 

For single-image tasks reported in Table~\ref{tab:main_single}, our models continue to manifest progress, presenting higher average scores. We improve the performance for two zero-shot VQA tasks and one MLLM benchmark, MMBench. However, our models only show modest performance on MME.  

\begin{table}[!t]
\centering
\resizebox{.48\textwidth}{!}{
\begin{tabular}{@{\hspace{0.05cm}}c@{\hspace{0.15cm}}c@{\hspace{0.15cm}}c@{\hspace{0.1cm}}c@{\hspace{0.1cm}}|@{\hspace{0.15cm}}l@{\hspace{0.2cm}}l@{\hspace{0.1cm}}} \toprule
Models & PT & FT & Drop & AVG-Multi & AVG \\ \midrule
InstructBLIP  & -  & - & -  & 42.56 & 50.34 \\
\midrule	
Ours-sampled  & \ding{56}  & \checkmark & \ding{56}  & 46.51 \color{blue}{(+3.96)} & 51.85 \color{blue}{(+1.51)} \\
Ours-sampled  & \checkmark & \checkmark & \ding{56}  & 47.12 \color{blue}{(+4.56)} & 50.94 \color{blue}{(+0.50)} \\
\noalign{\vskip 1pt} \cdashline{1-6}\noalign{\vskip 2pt}
Ours-sampled  & \checkmark & \checkmark & IMG-N & 48.06 & 52.09   \\
Ours-sampled  & \checkmark & \checkmark & IMG-B & 48.06  & 51.90  \\
Ours-sampled  & \checkmark & \checkmark & TXT & 48.08  & 52.08  \\ 
Ours-sampled  & \checkmark & \checkmark & ALL & \textbf{48.87} \color{blue}{(+6.31)} & \textbf{52.39} \color{blue}{(+2.05)} \\ \midrule
\textcolor{gray}{MMICL}  & \textcolor{gray}{-} & \textcolor{gray}{-} & \textcolor{gray}{-} & \textcolor{gray}{47.05} & \textcolor{gray}{51.68}\\ 
\bottomrule
\end{tabular}
}
\vspace{-0.5em}
\caption{Ablation study of different training strategies on XL-sized (3B LLM) models. ``PT'' refers to pre-training, and ``FT'' denotes fine-tuning. ``Ours-sampled'' is described in \S\ref{sec:abla_strategy}. ``AVG-Multi'' is the average score for multi-image settings, including A-OKVQA 4-shot, NLVR2, SEED video split and MSVD QA. ``AVG'' refers to the average score over 7 tasks, with detailed results presented in Appendix~\ref{appendix:abla}.} 
\vspace{-1em}
\label{tab:abla_strategy}
\end{table}

\subsection{Ablation Study}
\paragraph{Impact of condition context vectors.} \label{sec:abla_cond}
To determine whether condition context vectors $\mathcal{C}$ contribute to the improvement, we conduct ablation study by removing $\mathcal{C}$ and observe a decline in accuracy across various tasks, evaluation settings, and model scales. In detail, we replace these vectors by zero vectors to simulate the absence of $\mathcal{C}$. Experiments are conducted on zero-shot and few-shot VQA, and multi-image visual reasoning tasks. As shown in Table~\ref{tab:abla_cond}, the models augmented by $\mathcal{C}$ (``Ours'') consistently outperform those with zero vectors (``Ours-None''). The most substantial average discrepancy is observed in Brote-EX at the XL scale (2.03\%), while the smallest gap is presented by Brote-IM at the XXL scale (0.54\%). We notice that Brote-EX tends to gain more directly from $\mathcal{C}$ compared to Brote-IM, and conclude that Brote-IM directly integrates additional benefits provided by $\mathcal{C}$ into the model through dual-loss training. More sophisticated analysis are documented in \S\ref{sec:ex_vs_im}.

\paragraph{Impact of different training strategies.} \label{sec:abla_strategy}
For efficient iteration and validation, we create a subset by sampling one-third of the training data for ablation studies on different training strategies, denoting the resulting models as \textit{Ours-sampled}. For fair comparison, we also reproduce MMICL-XL (with InstructBLIP backbone) using this subset\footnote{We use the published code from \url{https://github.com/HaozheZhao/MIC}}. We evaluate the average scores of training strategies described in \S\ref{sec:fine_strategy} and \S\ref{sec:pre_strategy}, with a special focus on prior-LLM multimodal context fusion for multi-image scenarios. The averaged scores for multi-image tasks and the overall tasks are reported in Table~\ref{tab:abla_strategy}, with detailed results provided in Appendix~\ref{appendix:abla}. The performance of InstructBLIP serves as the baseline and is used to indicate the contribution of each of the designed strategies. Summarised from Table~\ref{tab:abla_strategy}, the models equipped with context-dropping strategies yield higher average scores. Notably, the dropping-ALL strategy presents the highest average scores of both multi-image and overall tasks, showing profound multimodal context handling ability. We consequently adopt this strategy for training our Brote models.

\begin{table}[t]
    \centering\small
    \begin{tabular}{@{\hspace{0.05cm}}l@{\hspace{0.1cm}}|@{\hspace{0.1cm}}c@{\hspace{0.1cm}}c@{\hspace{0.1cm}}c@{\hspace{0.1cm}}c@{\hspace{0.1cm}}|@{\hspace{0.1cm}}c@{\hspace{0.1cm}}c@{\hspace{0.05cm}}}
        \toprule
         Model & A-OK & NLVR2 & MSVD & SEED & AVG & Gain \\ \midrule
         Brote-EX & 56.00 & 71.41 & 53.02 & 57.51 & 59.49 & - \\
         \begin{tabular}[c]{@{}l@{}}Brote-EX\\\phantom{00}(\textit{+2epoch})\end{tabular} & 55.83 & 75.07 & 55.60 & 57.60 & 61.03 & 1.54\\
         \noalign{\vskip 1pt} \cdashline{1-7}\noalign{\vskip 2pt}
         Brote-IM & \textbf{56.53} & \textbf{76.02} & \textbf{56.06} & \textbf{57.86} & \textbf{61.62} & \textbf{2.13} \\
         \bottomrule
    \end{tabular}
    \caption{Results of continue training with XL models. ``Brote-EX (\textit{+2epoch})'' is training Brote-EX for 2 extra epochs using dual-loss without providing $\mathcal{C}$ for $\mathcal{M}_{C}$. ``A-OK'' is A-OKVQA for short. ``Gain'' implies the increment from extra epochs over original Brote-EX.}
    \label{tab:continu_training}
\end{table}

\section{Discussions and Analysis}

\subsection{Explicit Versus Implicit} \label{sec:ex_vs_im}
As discussed in \S\ref{sec:abla_cond}, Brote-EX exhibits a more significant benefit from $\mathcal{C}$ compared to Brote-IM. We propose two potential reasons for this observation: 
\begin{itemize}
    \setlength{\itemsep}{0pt}
    \setlength{\parsep}{0pt}
    \setlength{\parskip}{0pt}
    \item Brote-IM gains advantages from extra training steps rather than insights provided by $\mathcal{C}$;
    \item Brote-IM effectively incorporates the capabilities afforded by $\mathcal{C}$ into the parameters of LLM and Q-former during the training process.
\end{itemize}
For further investigation, we extend the training of Brote-EX with the same configurations and objectives as applied to Brote-IM, except that we zero out $\mathcal{C}$ for the concentrating phase. Specifically, we replace $\mathcal{C}$ in Equation \ref{eq:m_c} by zero vectors. Brote-IM is trained for two epochs based on Brote-EX as detailed in Appendix~\ref{appendix:training}. Hence, we train Brote-EX for additional two epochs, denoting this model as Brote-EX (\textit{+2epoch}). We report the results in Table~\ref{tab:continu_training}. The results reveal that the observed improvements of Brote-IM over Brote-EX are not solely attributable to increased training steps. Rather, the improvements stem from the integration with $\mathcal{C}$ during training. Although Brote-EX (\textit{+2epoch}) presents an average increase of 1.54\% over Brote-EX, Brote-IM exhibits an additional 0.59\% average improvement over Brote-EX (\textit{+2epoch}) with the participant of $\mathcal{C}$ during training, culminating in a total increment of 2.13\% over Brote-EX.

Furthermore, as detailed in Table~\ref{tab:abla_cond}, the absence of $\mathcal{C}$ does not prevent Brote-IM models from outperforming Brote-EX. This finding supports the conclusion that Brote-IM integrates the function of $\mathcal{C}$ into the model parameters themselves without explicitly generate $\mathcal{C}$ from the other model, facilitating a more profound comprehension of multimodal inputs. In contrast, Brote-EX relies on an extra explicit representation, condition context $\mathcal{C}$, to obtain multimodal comprehension and achieve good performances. The superior performance of Brote-IM affirms the efficacy of the dual-loss training strategy. Brote-IM markedly benefits from $\mathcal{C}$, thereby enabling the development of a more apt parameter set for multimodal context.

\subsection{Case Study}
\begin{figure}
    \centering
    \includegraphics[width=.50\textwidth]{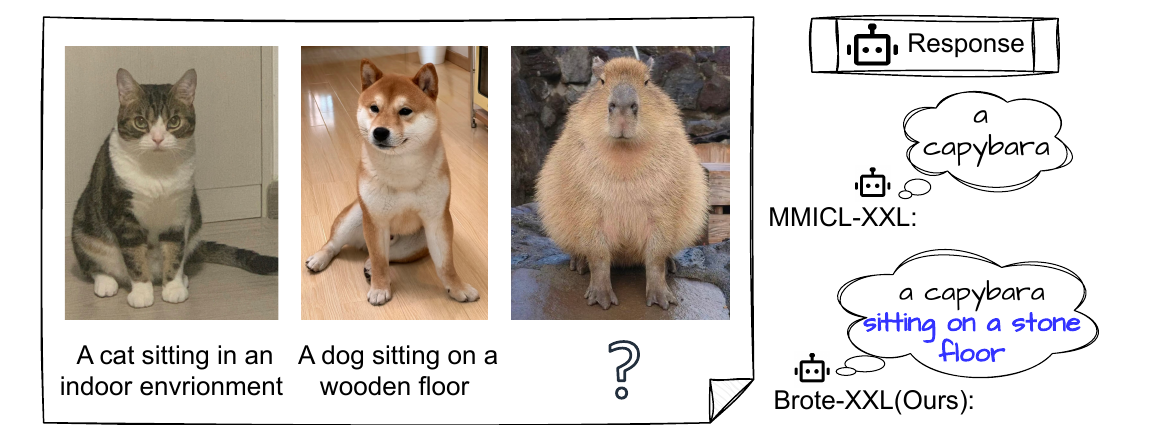}
    \vspace{-1.8em}
    \caption{A case showing that our method is more coherent to the given multimodal context.}
    \vspace{-1em}
    \label{fig:showcase}
\end{figure}

In Figure~\ref{fig:showcase}, we illustrate a case study on multimodal ICL, highlighting the coherent performance of our model in response to the input. Specifically, our model demonstrates an acute awareness of the target information conveyed through the multimodal inputs, capturing an intra-image connection characterized by ``an \textit{animal} sitting on/in \textit{a certain place}''. Compared to MMICL, our model produces a response that precisely aligns with the input, showcasing its profound ability to comprehend multimodal contexts.

\subsection{Efficiency of Inference}
We investigate the efficiency of inference with our models in terms of both time and GPU memory, as our methods involve two forward iterations and an additional $\mathcal{C}$ compared to other InstructBLIP-based models. We conduct experiments with XL models using single NVIDIA A100 GPU, with batch size 10 and data type float32. Results indicate that Brote-EX requires almost equal GPU memory (18G) and inference time (around 2.3 second per batch) compared to MMICL. However, Brote-IM exhibits an increase of GPU memory from 18G to 24G for an additional ``browsing'' iteration, and doubles the time cost to 5 second per batch. 

\section{Conclusion}
In this paper, we address the prior-LLM modality isolation issue for both image-text and inter-image context, which lacks sufficient investigation in previous works. To mitigate this issue, we propose browse-and-concentrate paradigm that leverages the initial browsing insights for the prior-LLM multimodal context fusion to stimulate more profound comprehending of multi-image and interleaved inputs. We present in-depth analysis on our proposed training strategies and the two approaches for implementing our proposed paradigm. The two approaches, explicitly or implicitly browse through and then concentrate on the context, exhibits comprehensive multimodal context understanding. Our method demonstrates remarkable improvements on 7 multi-image tasks against strong baselines that enable prior-LLM image-text fusion.

\section*{Limitations}
We conlude the limitations of our method as follows: First, although presenting improved results for multi-image scenarios, our method does not achieve equally impressive performances across all single-image tasks evaluated. This discrepancy can be attributed to the employed backbone models (InstructBLIP), which already incorporate the textual instructions into the visual feature extraction process, partially addressing the challenge of prior-LLM modality isolation we aim to overcome. Our future work includes validating the proposed paradigm on broader backbone models. Second, we do not specifically incorporate datasets designed for visual instruction tuning, such as LLaVA~\cite{liu2023visual}, which could be a reason for the modest performance on MME benchmark. In this paper, we primarily focuse on multi-image scenarios, such as question-answering and visual reasoning, without a particular emphasis on following visual instruction. Third, as we introduce a two-phase paradigm, the time cost and the required GPU memory for inference with Brote-IM are also increased. 


\section*{Acknowledgements}
This work is supported by the National Key R\&D Program of China (2022ZD0160502) and the National Natural Science Foundation of China (No. 61925601, 62276152). We appreciate all the reviewers for their insightful suggestions. We thank Siyu Wang for her participation in this work, and Haozhe Zhao for his technical support. We thank Tong Su for providing photos presented in Figure~\ref{fig:showcase}, and Alan (the cat) for being the model.

\section*{Contribution}

This work is contributed by students and supervisors from Tsinghua University. The following authors alone contributed to this paper and participated in every stage of this work, specifically:

\begin{itemize}[left=0.4cm, itemsep=2pt, parsep=0pt]
    \item Ziyue Wang: Paper composing, figure design, coding for training frameworks and experiments, data generation pipeline design. 
    \item Chi Chen: Paper composing, experiments design and support, training paradigm design. 
    \item Yiqi Zhu: Paper composing, data generation and processing, reproduction of baselines methods.
    \item Fuwen Luo: Paper composing, experiments design and support, training paradigm design.
    \item Peng Li: Supervision, paper composing, experiments design and support.
    \item Yang Liu: Supervision, paper composing, experiments design and support.
\end{itemize}

We would also be grateful to Professor Maosong Sun for leading the entire research group.

\bibliography{anthology,custom}

\begin{thebibliography}{44}
\expandafter\ifx\csname natexlab\endcsname\relax\def\natexlab#1{#1}\fi

\bibitem[{Alayrac et~al.(2022)Alayrac, Donahue, Luc, Miech, Barr, Hasson, Lenc, Mensch, Millican, Reynolds et~al.}]{alayrac2022flamingo}
Jean-Baptiste Alayrac, Jeff Donahue, Pauline Luc, Antoine Miech, Iain Barr, Yana Hasson, Karel Lenc, Arthur Mensch, Katherine Millican, Malcolm Reynolds, et~al. 2022.
\newblock Flamingo: a visual language model for few-shot learning.
\newblock \emph{Advances in Neural Information Processing Systems}, 35:23716--23736.

\bibitem[{Antol et~al.(2015)Antol, Agrawal, Lu, Mitchell, Batra, Zitnick, and Parikh}]{antol2015vqa}
Stanislaw Antol, Aishwarya Agrawal, Jiasen Lu, Margaret Mitchell, Dhruv Batra, C.~Lawrence Zitnick, and Devi Parikh. 2015.
\newblock \href {https://doi.org/10.1109/ICCV.2015.279} {{VQA:} visual question answering}.
\newblock In \emph{2015 {IEEE} International Conference on Computer Vision, {ICCV} 2015, Santiago, Chile, December 7-13, 2015}, pages 2425--2433. {IEEE} Computer Society.

\bibitem[{Biten et~al.(2019)Biten, Tito, Mafla, i~Bigorda, Rusi{\~{n}}ol, Jawahar, Valveny, and Karatzas}]{biten2019scene}
Ali~Furkan Biten, Rub{\`{e}}n Tito, Andr{\'{e}}s Mafla, Llu{\'{\i}}s~G{\'{o}}mez i~Bigorda, Mar{\c{c}}al Rusi{\~{n}}ol, C.~V. Jawahar, Ernest Valveny, and Dimosthenis Karatzas. 2019.
\newblock \href {https://doi.org/10.1109/ICCV.2019.00439} {Scene text visual question answering}.
\newblock In \emph{2019 {IEEE/CVF} International Conference on Computer Vision, {ICCV} 2019, Seoul, Korea (South), October 27 - November 2, 2019}, pages 4290--4300. {IEEE}.

\bibitem[{Chung et~al.(2022)Chung, Hou, Longpre, Zoph, Tay, Fedus, Li, Wang, Dehghani, Brahma et~al.}]{chung2022scaling}
Hyung~Won Chung, Le~Hou, Shayne Longpre, Barret Zoph, Yi~Tay, William Fedus, Yunxuan Li, Xuezhi Wang, Mostafa Dehghani, Siddhartha Brahma, et~al. 2022.
\newblock \href {https://arxiv.org/abs/2210.11416} {Scaling instruction-finetuned language models}.
\newblock \emph{ArXiv preprint}, abs/2210.11416.

\bibitem[{Dai et~al.(2023)Dai, Li, Li, Tiong, Zhao, Wang, Li, Fung, and Hoi}]{instructblip}
Wenliang Dai, Junnan Li, Dongxu Li, Anthony Meng~Huat Tiong, Junqi Zhao, Weisheng Wang, Boyang Li, Pascale Fung, and Steven Hoi. 2023.
\newblock \href {http://arxiv.org/abs/2305.06500} {Instruct{BLIP}: Towards general-purpose vision-language models with instruction tuning}.

\bibitem[{Dosovitskiy et~al.(2021)Dosovitskiy, Beyer, Kolesnikov, Weissenborn, Zhai, Unterthiner, Dehghani, Minderer, Heigold, Gelly, Uszkoreit, and Houlsby}]{dosovitskiy2020image}
Alexey Dosovitskiy, Lucas Beyer, Alexander Kolesnikov, Dirk Weissenborn, Xiaohua Zhai, Thomas Unterthiner, Mostafa Dehghani, Matthias Minderer, Georg Heigold, Sylvain Gelly, Jakob Uszkoreit, and Neil Houlsby. 2021.
\newblock \href {https://openreview.net/forum?id=YicbFdNTTy} {An image is worth 16x16 words: Transformers for image recognition at scale}.
\newblock In \emph{The Nineth International Conference on Learning Representations, {ICLR} 2021, Virtual Event, Austria, May 3-7, 2021}. OpenReview.net.

\bibitem[{Fu et~al.(2023)Fu, Chen, Shen, Qin, Zhang, Lin, Yang, Zheng, Li, Sun, Wu, and Ji}]{fu2023mme}
Chaoyou Fu, Peixian Chen, Yunhang Shen, Yulei Qin, Mengdan Zhang, Xu~Lin, Jinrui Yang, Xiawu Zheng, Ke~Li, Xing Sun, Yunsheng Wu, and Rongrong Ji. 2023.
\newblock \href {http://arxiv.org/abs/2306.13394} {{MME}: A comprehensive evaluation benchmark for multimodal large language models}.

\bibitem[{Garner(1987)}]{garner1987metacognition}
Ruth Garner. 1987.
\newblock \emph{Metacognition and reading comprehension.}
\newblock Ablex Publishing.

\bibitem[{Goyal et~al.(2017)Goyal, Khot, Summers{-}Stay, Batra, and Parikh}]{goyal2017making}
Yash Goyal, Tejas Khot, Douglas Summers{-}Stay, Dhruv Batra, and Devi Parikh. 2017.
\newblock \href {https://doi.org/10.1109/CVPR.2017.670} {Making the {V} in {VQA} matter: Elevating the role of image understanding in visual question answering}.
\newblock In \emph{2017 {IEEE} Conference on Computer Vision and Pattern Recognition, {CVPR} 2017, Honolulu, HI, USA, July 21-26, 2017}, pages 6325--6334. {IEEE} Computer Society.

\bibitem[{Hu et~al.(2022{\natexlab{a}})Hu, Shen, Wallis, Allen{-}Zhu, Li, Wang, Wang, and Chen}]{hu2021lora}
Edward~J. Hu, Yelong Shen, Phillip Wallis, Zeyuan Allen{-}Zhu, Yuanzhi Li, Shean Wang, Lu~Wang, and Weizhu Chen. 2022{\natexlab{a}}.
\newblock \href {https://openreview.net/forum?id=nZeVKeeFYf9} {Lora: Low-rank adaptation of large language models}.
\newblock In \emph{The Tenth International Conference on Learning Representations, {ICLR} 2022, Virtual Event, April 25-29, 2022}. OpenReview.net.

\bibitem[{Hu et~al.(2022{\natexlab{b}})Hu, Hua, Yang, Shi, Smith, and Luo}]{hu2023promptcap}
Yushi Hu, Hang Hua, Zhengyuan Yang, Weijia Shi, Noah~A Smith, and Jiebo Luo. 2022{\natexlab{b}}.
\newblock \href {https://arxiv.org/abs/2211.09699} {Prompt{C}ap: Prompt-guided task-aware image captioning}.
\newblock \emph{ArXiv preprint}, abs/2211.09699.

\bibitem[{Huang et~al.(2023)Huang, Dong, Wang, Hao, Singhal, Ma, Lv, Cui, Mohammed, Patra, Liu, Aggarwal, Chi, Bjorck, Chaudhary, Som, Song, and Wei}]{huang2023language}
Shaohan Huang, Li~Dong, Wenhui Wang, Yaru Hao, Saksham Singhal, Shuming Ma, Tengchao Lv, Lei Cui, Owais~Khan Mohammed, Barun Patra, Qiang Liu, Kriti Aggarwal, Zewen Chi, Johan Bjorck, Vishrav Chaudhary, Subhojit Som, Xia Song, and Furu Wei. 2023.
\newblock \href {http://arxiv.org/abs/2302.14045} {Language is not all you need: Aligning perception with language models}.

\bibitem[{Li et~al.(2023{\natexlab{a}})Li, Zhang, Chen, Wang, Pu, Yang, Li, and Liu}]{li2023mimic}
Bo~Li, Yuanhan Zhang, Liangyu Chen, Jinghao Wang, Fanyi Pu, Jingkang Yang, Chunyuan Li, and Ziwei Liu. 2023{\natexlab{a}}.
\newblock \href {https://arxiv.org/abs/2306.05425} {Mimic-it: Multi-modal in-context instruction tuning}.
\newblock \emph{ArXiv preprint}, abs/2306.05425.

\bibitem[{Li et~al.(2023{\natexlab{b}})Li, Zhang, Chen, Wang, Yang, and Liu}]{li2023otter}
Bo~Li, Yuanhan Zhang, Liangyu Chen, Jinghao Wang, Jingkang Yang, and Ziwei Liu. 2023{\natexlab{b}}.
\newblock \href {https://arxiv.org/abs/2305.03726} {Otter: A multi-modal model with in-context instruction tuning}.
\newblock \emph{ArXiv preprint}, abs/2305.03726.

\bibitem[{Li et~al.(2023{\natexlab{c}})Li, Wang, Wang, Ge, Ge, and Shan}]{li2023seedbench}
Bohao Li, Rui Wang, Guangzhi Wang, Yuying Ge, Yixiao Ge, and Ying Shan. 2023{\natexlab{c}}.
\newblock \href {http://arxiv.org/abs/2307.16125} {{SEED-B}ench: Benchmarking multimodal llms with generative comprehension}.

\bibitem[{Li et~al.(2023{\natexlab{d}})Li, Pan, Ge, Gao, Zhang, Ji, Zhang, Chua, Tang, and Zhuang}]{li2023fine}
Juncheng Li, Kaihang Pan, Zhiqi Ge, Minghe Gao, Hanwang Zhang, Wei Ji, Wenqiao Zhang, Tat-Seng Chua, Siliang Tang, and Yueting Zhuang. 2023{\natexlab{d}}.
\newblock \href {https://arxiv.org/abs/2308.04152} {Fine-tuning multimodal llms to follow zeroshot demonstrative instructions}.
\newblock \emph{ArXiv preprint}, abs/2308.04152.

\bibitem[{Li et~al.(2023{\natexlab{e}})Li, Li, Savarese, and Hoi}]{li2023blip}
Junnan Li, Dongxu Li, Silvio Savarese, and Steven Hoi. 2023{\natexlab{e}}.
\newblock \href {https://arxiv.org/abs/2301.12597} {{BLIP}-2: Bootstrapping language-image pre-training with frozen image encoders and large language models}.
\newblock \emph{ArXiv preprint}, abs/2301.12597.

\bibitem[{Liu et~al.(2023{\natexlab{a}})Liu, Emerson, and Collier}]{liu2023vsr}
Fangyu Liu, Guy Emerson, and Nigel Collier. 2023{\natexlab{a}}.
\newblock Visual spatial reasoning.
\newblock \emph{Transactions of the Association for Computational Linguistics}, 11:635--651.

\bibitem[{Liu et~al.(2023{\natexlab{b}})Liu, Zhu, Wu, Yang, You, Wang, Lu, Liu, Zheng, Sun et~al.}]{liu2023medical}
Fenglin Liu, Tingting Zhu, Xian Wu, Bang Yang, Chenyu You, Chenyang Wang, Lei Lu, Zhangdaihong Liu, Yefeng Zheng, Xu~Sun, et~al. 2023{\natexlab{b}}.
\newblock A medical multimodal large language model for future pandemics.
\newblock \emph{NPJ Digital Medicine}, 6(1):226.

\bibitem[{Liu et~al.(2023{\natexlab{c}})Liu, Li, Li, and Lee}]{liu2023improved}
Haotian Liu, Chunyuan Li, Yuheng Li, and Yong~Jae Lee. 2023{\natexlab{c}}.
\newblock \href {https://arxiv.org/abs/2310.03744} {Improved baselines with visual instruction tuning}.
\newblock \emph{ArXiv preprint}, abs/2310.03744.

\bibitem[{Liu et~al.(2023{\natexlab{d}})Liu, Li, Wu, and Lee}]{liu2023visual}
Haotian Liu, Chunyuan Li, Qingyang Wu, and Yong~Jae Lee. 2023{\natexlab{d}}.
\newblock \href {https://arxiv.org/abs/2304.08485} {Visual instruction tuning}.
\newblock \emph{ArXiv preprint}, abs/2304.08485.

\bibitem[{Liu et~al.(2023{\natexlab{e}})Liu, Duan, Zhang, Li, Zhang, Zhao, Yuan, Wang, He, Liu, Chen, and Lin}]{liu2023mmbench}
Yuan Liu, Haodong Duan, Yuanhan Zhang, Bo~Li, Songyang Zhang, Wangbo Zhao, Yike Yuan, Jiaqi Wang, Conghui He, Ziwei Liu, Kai Chen, and Dahua Lin. 2023{\natexlab{e}}.
\newblock \href {http://arxiv.org/abs/2307.06281} {Mmbench: Is your multi-modal model an all-around player?}

\bibitem[{Lu et~al.(2021)Lu, Qiu, Chen, Xia, Zhao, Zhang, Yu, Liang, and Zhu}]{lu2021iconqa}
Pan Lu, Liang Qiu, Jiaqi Chen, Tony Xia, Yizhou Zhao, Wei Zhang, Zhou Yu, Xiaodan Liang, and Song-Chun Zhu. 2021.
\newblock Iconqa: A new benchmark for abstract diagram understanding and visual language reasoning.
\newblock In \emph{The 35th Conference on Neural Information Processing Systems (NeurIPS) Track on Datasets and Benchmarks}.

\bibitem[{Luo et~al.(2023)Luo, Zhou, Ren, Chen, Sun, and Ji}]{luo2023cheap}
Gen Luo, Yiyi Zhou, Tianhe Ren, Shengxin Chen, Xiaoshuai Sun, and Rongrong Ji. 2023.
\newblock \href {https://arxiv.org/abs/2305.15023} {Cheap and quick: Efficient vision-language instruction tuning for large language models}.
\newblock \emph{ArXiv preprint}, abs/2305.15023.

\bibitem[{OpenAI(2023)}]{OpenAI2023Gpt4v}
OpenAI. 2023.
\newblock \href {https://arxiv.org/abs/2303.08774} {Gpt-4 technical report.}
\newblock \emph{ArXiv preprint}, abs/2303.08774.

\bibitem[{Qi et~al.(2023)Qi, Fang, Zhang, Sun, Wu, Liu, Lin, Wang, and Zhao}]{qi2023gemini}
Zhangyang Qi, Ye~Fang, Mengchen Zhang, Zeyi Sun, Tong Wu, Ziwei Liu, Dahua Lin, Jiaqi Wang, and Hengshuang Zhao. 2023.
\newblock \href {http://arxiv.org/abs/2312.15011} {Gemini vs {GPT-4V}: A preliminary comparison and combination of vision-language models through qualitative cases}.

\bibitem[{Radford et~al.(2021)Radford, Kim, Hallacy, Ramesh, Goh, Agarwal, Sastry, Askell, Mishkin, Clark, Krueger, and Sutskever}]{radford2021learning}
Alec Radford, Jong~Wook Kim, Chris Hallacy, Aditya Ramesh, Gabriel Goh, Sandhini Agarwal, Girish Sastry, Amanda Askell, Pamela Mishkin, Jack Clark, Gretchen Krueger, and Ilya Sutskever. 2021.
\newblock \href {http://proceedings.mlr.press/v139/radford21a.html} {Learning transferable visual models from natural language supervision}.
\newblock In \emph{Proceedings of the 38th International Conference on Machine Learning, {ICML} 2021, 18-24 July 2021, Virtual Event}, volume 139 of \emph{Proceedings of Machine Learning Research}, pages 8748--8763. {PMLR}.

\bibitem[{Saikh et~al.(2022)Saikh, Ghosal, Mittal, Ekbal, and Bhattacharyya}]{saikh2022scienceqa}
Tanik Saikh, Tirthankar Ghosal, Amish Mittal, Asif Ekbal, and Pushpak Bhattacharyya. 2022.
\newblock Science{QA}: a novel resource for question answering on scholarly articles.
\newblock \emph{International Journal on Digital Libraries}, 23(3):289--301.

\bibitem[{Schwenk et~al.(2022)Schwenk, Khandelwal, Clark, Marino, and Mottaghi}]{schwenk2022aok}
Dustin Schwenk, Apoorv Khandelwal, Christopher Clark, Kenneth Marino, and Roozbeh Mottaghi. 2022.
\newblock {A-OKVQA:} a benchmark for visual question answering using world knowledge.
\newblock In \emph{Computer Vision--ECCV 2022: 17th European Conference, Tel Aviv, Israel, October 23--27, 2022, Proceedings, Part VIII}, pages 146--162. Springer.

\bibitem[{Shukor et~al.(2023)Shukor, Rame, Dancette, and Cord}]{shukor2023beyond}
Mustafa Shukor, Alexandre Rame, Corentin Dancette, and Matthieu Cord. 2023.
\newblock \href {https://arxiv.org/abs/2310.00647} {Beyond task performance: Evaluating and reducing the flaws of large multimodal models with in-context learning}.
\newblock \emph{ArXiv preprint}, abs/2310.00647.

\bibitem[{Suhr et~al.(2019)Suhr, Zhou, Zhang, Zhang, Bai, and Artzi}]{suhr2019corpus}
Alane Suhr, Stephanie Zhou, Ally Zhang, Iris Zhang, Huajun Bai, and Yoav Artzi. 2019.
\newblock \href {https://doi.org/10.18653/v1/P19-1644} {A corpus for reasoning about natural language grounded in photographs}.
\newblock In \emph{Proceedings of the 57th Annual Meeting of the Association for Computational Linguistics}, pages 6418--6428, Florence, Italy. Association for Computational Linguistics.

\bibitem[{Sun et~al.(2023)Sun, Cui, Zhang, Zhang, Yu, Luo, Wang, Rao, Liu, Huang, and Wang}]{sun2023generative}
Quan Sun, Yufeng Cui, Xiaosong Zhang, Fan Zhang, Qiying Yu, Zhengxiong Luo, Yueze Wang, Yongming Rao, Jingjing Liu, Tiejun Huang, and Xinlong Wang. 2023.
\newblock \href {http://arxiv.org/abs/2312.13286} {Generative multimodal models are in-context learners}.

\bibitem[{Team et~al.(2023)Team, Anil, Borgeaud, Wu, Alayrac, Yu, Soricut, Schalkwyk, Dai, Hauth et~al.}]{team2023gemini}
Gemini Team, Rohan Anil, Sebastian Borgeaud, Yonghui Wu, Jean-Baptiste Alayrac, Jiahui Yu, Radu Soricut, Johan Schalkwyk, Andrew~M Dai, Anja Hauth, et~al. 2023.
\newblock \href {https://arxiv.org/abs/2312.11805} {Gemini: a family of highly capable multimodal models}.
\newblock \emph{ArXiv preprint}, abs/2312.11805.

\bibitem[{Tsimpoukelli et~al.(2021)Tsimpoukelli, Menick, Cabi, Eslami, Vinyals, and Hill}]{tsimpoukelli2021multimodal}
Maria Tsimpoukelli, Jacob Menick, Serkan Cabi, S.~M.~Ali Eslami, Oriol Vinyals, and Felix Hill. 2021.
\newblock \href {https://proceedings.neurips.cc/paper/2021/hash/01b7575c38dac42f3cfb7d500438b875-Abstract.html} {Multimodal few-shot learning with frozen language models}.
\newblock In \emph{Advances in Neural Information Processing Systems 34: Annual Conference on Neural Information Processing Systems 2021, NeurIPS 2021, December 6-14, 2021, virtual}, pages 200--212.

\bibitem[{Wang et~al.(2023)Wang, Lv, Yu, Hong, Qi, Wang, Ji, Yang, Zhao, Song, Xu, Xu, Li, Dong, Ding, and Tang}]{wang2023cogvlm}
Weihan Wang, Qingsong Lv, Wenmeng Yu, Wenyi Hong, Ji~Qi, Yan Wang, Junhui Ji, Zhuoyi Yang, Lei Zhao, Xixuan Song, Jiazheng Xu, Bin Xu, Juanzi Li, Yuxiao Dong, Ming Ding, and Jie Tang. 2023.
\newblock \href {http://arxiv.org/abs/2311.03079} {Cog{VLM}: Visual expert for pretrained language models}.

\bibitem[{Wei et~al.(2022)Wei, Wang, Schuurmans, Bosma, Xia, Chi, Le, Zhou et~al.}]{wei2022chain}
Jason Wei, Xuezhi Wang, Dale Schuurmans, Maarten Bosma, Fei Xia, Ed~H Chi, Quoc~V Le, Denny Zhou, et~al. 2022.
\newblock Chain-of-{T}hought prompting elicits reasoning in large language models.
\newblock In \emph{Advances in Neural Information Processing Systems}.

\bibitem[{Wu et~al.(2023)Wu, Gan, Chen, Wan, and Yu}]{wu2023multimodal}
Jiayang Wu, Wensheng Gan, Zefeng Chen, Shicheng Wan, and Philip~S. Yu. 2023.
\newblock \href {http://arxiv.org/abs/2311.13165} {Multimodal large language models: A survey}.

\bibitem[{Xu et~al.(2017)Xu, Zhao, Xiao, Wu, Zhang, He, and Zhuang}]{xu2017video}
Dejing Xu, Zhou Zhao, Jun Xiao, Fei Wu, Hanwang Zhang, Xiangnan He, and Yueting Zhuang. 2017.
\newblock \href {https://doi.org/10.1145/3123266.3123427} {Video question answering via gradually refined attention over appearance and motion}.
\newblock In \emph{Proceedings of the 2017 {ACM} on Multimedia Conference, {MM} 2017, Mountain View, CA, USA, October 23-27, 2017}, pages 1645--1653.

\bibitem[{Yang et~al.(2021)Yang, Miech, Sivic, Laptev, and Schmid}]{yang2021justask}
Antoine Yang, Antoine Miech, Josef Sivic, Ivan Laptev, and Cordelia Schmid. 2021.
\newblock \href {https://doi.org/10.1109/ICCV48922.2021.00171} {Just ask: Learning to answer questions from millions of narrated videos}.
\newblock In \emph{2021 {IEEE/CVF} International Conference on Computer Vision, {ICCV} 2021, Montreal, QC, Canada, October 10-17, 2021}, pages 1666--1677. {IEEE}.

\bibitem[{Ye et~al.(2023)Ye, Xu, Xu, Ye, Yan, Zhou, Wang, Hu, Shi, Shi et~al.}]{ye2023mplug}
Qinghao Ye, Haiyang Xu, Guohai Xu, Jiabo Ye, Ming Yan, Yiyang Zhou, Junyang Wang, Anwen Hu, Pengcheng Shi, Yaya Shi, et~al. 2023.
\newblock \href {https://arxiv.org/abs/2304.14178} {mplug-owl: Modularization empowers large language models with multimodality}.
\newblock \emph{ArXiv preprint}, abs/2304.14178.

\bibitem[{Yin et~al.(2023)Yin, Fu, Zhao, Li, Sun, Xu, and Chen}]{yin2023survey}
Shukang Yin, Chaoyou Fu, Sirui Zhao, Ke~Li, Xing Sun, Tong Xu, and Enhong Chen. 2023.
\newblock \href {https://arxiv.org/abs/2306.13549} {A survey on multimodal large language models}.
\newblock \emph{ArXiv preprint}, abs/2306.13549.

\bibitem[{Zellers et~al.(2019)Zellers, Bisk, Farhadi, and Choi}]{zellers2019vcr}
Rowan Zellers, Yonatan Bisk, Ali Farhadi, and Yejin Choi. 2019.
\newblock \href {https://doi.org/10.1109/CVPR.2019.00688} {From recognition to cognition: Visual commonsense reasoning}.
\newblock In \emph{2019 {IEEE} Conference on Computer Vision and Pattern Recognition, {CVPR} 2019, Long Beach, CA, USA, June 16-20, 2019}, pages 6720--6731. Computer Vision Foundation / {IEEE}.

\bibitem[{Zhao et~al.(2024)Zhao, Cai, Si, Ma, An, Chen, Liu, Wang, Han, and Chang}]{zhao2023mmicl}
Haozhe Zhao, Zefan Cai, Shuzheng Si, Xiaojian Ma, Kaikai An, Liang Chen, Zixuan Liu, Sheng Wang, Wenjuan Han, and Baobao Chang. 2024.
\newblock \href {https://openreview.net/forum?id=5KojubHBr8} {{MMICL}: Empowering vision-language model with multi-modal in-context learning}.
\newblock In \emph{The Twelfth International Conference on Learning Representations}.

\bibitem[{Zhu et~al.(2023)Zhu, Chen, Shen, Li, and Elhoseiny}]{zhu2023minigpt}
Deyao Zhu, Jun Chen, Xiaoqian Shen, Xiang Li, and Mohamed Elhoseiny. 2023.
\newblock \href {https://arxiv.org/abs/2304.10592} {{MIMIGPT}-4: Enhancing vision-language understanding with advanced large language models}.
\newblock \emph{ArXiv preprint}, abs/2304.10592.

\end{thebibliography}
\bibliographystyle{acl_natbib}

\vspace{1em}
\appendix

\section{Details of Our Constructed Data for Pre-training}\label{appendix:stage1data}
In this section, we provide details of the data generation process mentioned in \S\ref{sec:data}.
Inspired by PromptCap~\cite{hu2023promptcap}, we employ LLM APIs to generate target-aware image descriptions, but explore broader types of tasks. Extending from single-image descriptions, we require the LLM\footnote{We use the GPT-4 API with version ``gpt-4-1106-preview''} to generate descriptions regarding other images as well, enabling a more profound understanding of multi-image and interleaved content. We utilize datasets targeting difference aspects of visual reasoning, including the maintenance of general world knowledge, the spatial and temporal information, the OCR ability, and the ability to distinguish differences of images. These datasets are as follows: 

\begin{itemize}[left=0.4cm, itemsep=2pt, parsep=0pt]
    \item \textbf{VQAv2}~\cite{goyal2017making} is a single-image visual captioning dataset. It is utilized to consolidate the general ability of our model.
    \item \textbf{ST-VQA}~\cite{biten2019scene} and \textbf{IconQA}~\cite{lu2021iconqa} are two single-image VQA datasets. They primarily conrtibute the tasks of OCR, object identification and counting. 
    \item \textbf{VSR}~\cite{liu2023vsr} is a VQA dataset addressing the spatial relation between two objects. It is employed to enhance the spatial reasoning ability.  
    \item \textbf{VCR}~\cite{zellers2019vcr} is a visual reasoning dataset with images extracted from video scenes, focusing on the relations between presenting figures and objects.
    \item \textbf{NLVR2}~\cite{suhr2019corpus} and \textbf{MIMIC-IT(CGD)}~\cite{li2023otter} both contain two interleaved images with text. They help with the ability to distinguish the differences between two images. 
    \item \textbf{iVQA}~\cite{yang2021justask} is a video question answering dataset. We utilize this dataset to promote the ability to deal with the sequential information.
\end{itemize}
\vspace{-6pt}
For datasets with naturally interleaved formats, VCR, NLVR2, and MIMIC-IT, we directly employ them to prompt LLMs, aiming to generate task-specific and multi-image aware descriptions. For other datasets, VQAv2, ST-VQA, IconQA, VSR, and iVQA, we adopt their few-shot versions from the MIC dataset~\cite{zhao2023mmicl}. In these versions, single-image instances are reconfigured into a few-shot format, featuring one or multiple images for zero to eight shots. These adapted instances serve as $QA^{J}$, as detailed in \S\ref{sec:data}, with the corresponding questions designed as targeted tasks for LLM responses. The statistics of our generated data are listed in Table~\ref{tab:data_gen_detail}. We split instances containing multiple images into single image paired with the corresponding descriptions, and use them for pre-training as described in \S\ref{sec:pre_strategy}.

\begin{table}[t]
\centering\small
\begin{tabular}{l@{\hspace{0.2cm}}c@{\hspace{0.2cm}}c@{\hspace{0.2cm}}c@{\hspace{0.1cm}}c}
\toprule
\begin{tabular}[c]{@{}c@{}}Original\\Dataset\end{tabular}  & Task Type & Format & \begin{tabular}[c]{@{}c@{}}\#Generated\\Data\end{tabular}  & \begin{tabular}[c]{@{}c@{}}\#Training \\Pairs \end{tabular}\\ \midrule
$^{\diamondsuit}$IconQA & VQA & S + M & 1.8k & 5.4k \\
$^{\heartsuit}$VSR & VQA & S + M & 3.0k & 14.4k \\ 
$^{\diamondsuit}$VQAv2 & VQA & S + M & 6.5k & 19.3k \\ 
$^{\triangle}$STVQA & VQA & S + M & 10.0k & 27.2k \\
$^{\square}$NLVR2 & Reasoning & I & 10.0k & 20.0k \\ 
$^{\spadesuit}$CGD* & Reasoning & I & 10.2k & 20.5k \\ 
$^{\clubsuit}$VCR & Reasoning & I & 10.0k & 62.0k \\ 
$^{\triangle}$iVQA & Video-QA & I & 5.2k & 22.4k \\ \midrule
\multicolumn{3}{c}{Total} & 56.7k & 191.2k \\ 
\bottomrule
\end{tabular}
\vspace{-6pt}
\caption{The statistics of our pre-training data. ``S'': \textbf{single}-image input; ``M'': \textbf{multi}-image input, wherein this case, the multiple images come from the few-shot examples of sinlge-image QA pairs; ``I'': naturally \textbf{interleaved} data, such as VCR and videos. ``CGD*'': MIMIC-IT CGD task~\cite{li2023mimic}. $^{\diamondsuit}$: datasets licensed under CC-BY 4.0. $^{\heartsuit}$: datasets licensed under Apache License, Version 2.0. $^{\spadesuit}$: datasets licensed under MIT License. $^{\clubsuit}$: datasets licensed under Custom License. $^{\triangle}$: datasets with unknown license. $^{\square}$: datasets with CC-BY 4.0 License for annotations and unknown license for images.} \label{tab:data_gen_detail}
\label{tab:stage1data}
\end{table}

\subsection{Prompt Templates for Pre-training Data Generation}
To be consistent with \S\ref{sec:data}, we represent our employed prompt including task instruction $\mathcal{P}$, the general image descriptions $C^{K}$, and the question-answer pairs $QA^{J}$, where $K$ and $J$ are the number of images and targeted question-answer pairs. We provide prompt templates for the included datasets as belows:

\begin{tcolorbox}[title=\small Prompt Template for Data Generation]
    \small 
    \texttt{General task instruction $\mathcal{P}$} \\
    \texttt{========} \\
    \texttt{The general descriptions <caption> for each image are as follows:} \\
    \texttt{$C^{1}$, $C^{2}$, $\dots$, $C^{K}$} \\
    \texttt{========} \\
    \texttt{Here are the additional information of <Question>-<Answer> you should focus on!} \\
    \texttt{$QA^{1}$, $QA^{2}$, $\dots$, $QA^{J}$} 
\end{tcolorbox}

Detailed task instruction of $\mathcal{P}$ for different datasets are as follows:
\begin{itemize} 
    \item Task Instruction for VQAv2 \& VCR \& IconQA \& ST-VQA \& VSR: \newline\textit{Generate detailed captions of each image involved in the following text according to the original caption and the given <Question>-<Answer> pairs. You should pay attention to the information in the <Answer>. Your output should be in the json format, as \{"image0":"", "image1":"", "image2":""\}. Your output should also be natural as an original caption and not include words like "answer" or "caption"!}
    \item Task Instruction for NLVR2: \newline\textit{Generate detailed captions of each image involved in the following text according to the original caption and the given <Question>-<Answer> pairs. You should pay attention to the information in the <Answer>. Your output should be in the json format, as \{"image0":"", "image1":"", "image2":""\}. Your output should also be natural as an original caption and not include words like "answer" or "caption"! You should also notice that <image0> is the left image and <image1> is the right image.}
    \item Task Instruction for iVQA: \newline\textit{Generate detailed captions of each image involved in the following text according to the original caption and the given <Question>-<Answer> pairs. You should pay attention to the information in the <Answer>. Your output should be in the json format, as \{"image0":"", "image1":"", "image2":""\}. Your output should also be natural as an original caption and not include words like "answer" or "caption"! You should notice that there exists sequential information between images!}
    \item Task Instruction for Task Instruction for MIMIC-IT(CGD): \newline\textit{Generate detailed captions of each image involved in the following text according to the original caption and the given <Option>-<Answer> pairs. You should pay attention to the information in the <Answer>. Your output should be in the json format, as \{"image0":"", "image1":"", "image2":""\}. Your output should also be natural as an original caption and not include words like "answer" or "caption"! Your output should also not clearly contain comparison while the information in <Option>-<Answer> pair should be presented!}
\end{itemize}

Here is a detailed example from ST-VQA:

\begin{tcolorbox}[title=An Example for Data Generation]
    \small 
    \texttt{Generate detailed captions of each image involved in the following text according to the original caption and the given <Question>-<Answer> pairs. You should pay attention to the information in the <Answer>. Your output should be in the json format, as \{"image0":"", "image1":"", "image2":""\}. Your output should also be as natural as an original caption and not include words like "answer" or "caption"!} \\
    \texttt{========} \\
    \texttt{The original caption for each image are as follows.} \\
    \texttt{<image0>: a sign with chinese characters on it;} \\
    \texttt{<image1>: a man is walking down a hallway with a television above him.} \\
    \texttt{========} \\
    \texttt{Here are the additional information that you should focus on!} \\ \\
    \texttt{The image 0: <image0> is the primary source of information for answering the questions. Please refer to it carefully when answering question: What does the street sign say? Answer: anping jie} \\ \\
    \texttt{Answer each question based on the information presented in image 1: <image1>. Given the picture <image1>, what is the answer to the question: What does the green sign say? Answer: exit} \\ \\
\end{tcolorbox}
The corresponding outputs for the two images are ``The green street sign displays the words `anping jie' in Chinese characters.'' and ``A man is strolling through a hallway while a television monitor is mounted above him alongside an indication of an `exit' on a green sign.''

\section{Training data for Model Fine-tuning}\label{appendix:trainingdata}
The select 17 datasets targeting different tasks from MIC dataset. Following \citet{instructblip} and \citet{zhao2023mmicl}, We sampled about 490k instances from MIC according to this equation:
\begin{equation}
    p_{d}=\frac{\sqrt{N_{d}}}{{\textstyle \sum_{D}^{i=1}} \sqrt{N_{i}}}, 
\end{equation}
where $p_{d}$ refers to the probability to select N instances of dataset $d$, from a total of D datasets. We list the involved datasets in Table~\ref{tab:stage2data}.

\begin{table}[t]
\centering\small
\begin{tabular}{@{\hspace{0.05cm}}l@{\hspace{0.1cm}}cc}
\toprule
Dataset & Task & Format \\ \midrule
$^{\spadesuit}$COCO & Captioning & paired \& few-shot \\
$^{\triangle}$Flickr & Captioning & paired \& few-shot \\
$^{\triangle}$MSRVTT & Captioning & interleaved \\
$^{\heartsuit}$VSR & Visual Reasoning & paired \& few-shot \\
$^{\square}$NLVR2 & Visual Reasoning & interleaved \\
$^{\spadesuit}$VCR & Visual Reasoning & interleaved \\
$^{\triangle}$OKVQA & VQA & paired \& few-shot \\
$^{\diamondsuit}$VQAv2 & VQA & paired \& few-shot \\
$^{\triangle}$GQA & VQA & paired \& few-shot \\
$^{\triangle}$STVQA & VQA & paired \& few-shot \\
$^{\diamondsuit}$TextVQA & VQA & paired \& few-shot \\
$^{\heartsuit}$RefCOCO & VQA & paired \& few-shot \\
$^{\clubsuit}$WikiART & VQA & paired \& few-shot \\
$^{\diamondsuit}$IconQA & VQA & paired \& few-shot \\
$^{\triangle}$iVQA & Video QA & interleaved \\
$^{\triangle}$MSVD & Video QA & interleaved \\
$^{\triangle}$MiniImageNet & Classification & paired \& few-shot \\
\bottomrule
\end{tabular}
\vspace{-1em}
\caption{An overview of our fine-tuning data. $^{\diamondsuit}$: datasets licensed under CC-BY 4.0. $^{\heartsuit}$: datasets licensed under Apache License, Version 2.0. $^{\spadesuit}$: datasets licensed under Custom License. $^{\clubsuit}$: datasets licensed under Non-commercial. $^{\triangle}$: datasets with unknown license. $^{\square}$: datasets with CC-BY 4.0 License for annotations and unknown license for images.} 
\vspace{-1em}
\label{tab:stage2data}
\end{table}

\section{Model Training}\label{appendix:training}
This section describes detailed pre-training and fine-tune setting. 

\subsection{Pre-training}
\begin{figure*}[t] 
\begin{center}
\includegraphics[width=\textwidth]{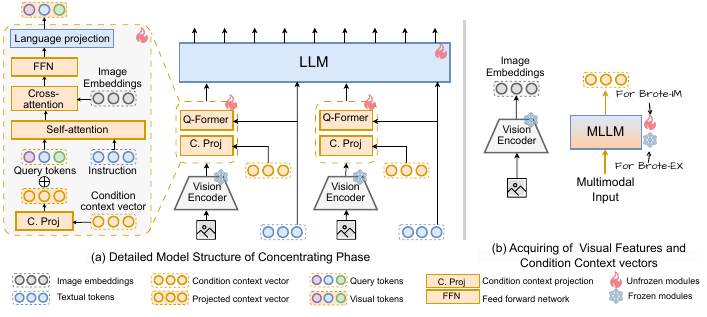}
\caption{The detailed model structure for concentrating phase.} \label{fig:modelv1}
\end{center}
\end{figure*}

We initially acquire all the condition context vectors for the data outlined in Appendix~\ref{appendix:stage1data} by MMICL models, where MMICL-XL and MMICL-XXL models are employed for Brote-XL and Brote-XXL, respectively. Leveraging these vectors, we bypass the forward iteration stage during pre-training and directly proceed the concentrating phase. We set the learning rate for the condition projection at $1 \times 10^{-4}$ and for both the Q-Former and the language projection at $1 \times 10^{-5}$, applying a cosine learning rate scheduler. These experiments are conducted on the NVIDIA A100 GPU, with the pre-training configurations detailed in Table~\ref{tab:pre_setting}. As a complement to Figure~\ref{fig:model}, we provide a detailed models structure in Figure~\ref{fig:modelv1}.

\subsection{Fine-tuning}
In the pre-training stage, we adapt the parameters of the Q-Former and the condition projection to effectively integrate $\mathcal{C}$, enhancing the models' ability to interpret multimodal contexts. Based on this, the subsequent fine-tuning encompasses both browsing and concentrating phases. Following \citet{instructblip} and \citet{zhao2023mmicl}, we fine-tune our model on multiple originated datasets to enable the ability to accomplish practical and diverse tasks.
As described in \S\ref{sec:overview}, we develop two approached for incorporating $\mathcal{C}$, each predicated on differing objectives: For Brote-EX, condition context vectors are derived from the frozen MMICL model, whereas Brote-IM generates these vectors internally. Training specifics for Brote-EX-XL involve four epochs focused on the objective $\mathcal{L}_{\mathcal{M}_{C}}$, while Brote-IM-XL extends this with two epochs under a dual-loss objective, $\mathcal{L}_{\mathcal{M}_{B}} + \mathcal{L}_{\mathcal{M}_{C}}$, starting from the Brote-EX-XL foundation. For the XXL models, the training duration for both explicit and implicit training modes are adjusted to half that of their XL counterparts. Detailed configurations are listed in Table~\ref{tab:fine_setting}, not that the settings of learning rates identical to that of pre-training stage.

\begin{table}[t]
\centering\small
\begin{tabular}{lc@{\hspace{0.15cm}}c@{\hspace{0.15cm}}c@{\hspace{0.15cm}}c@{\hspace{0.15cm}}c}
\toprule
\begin{tabular}[c]{@{}c@{}}Model\\Scale\end{tabular} & Epoch & \begin{tabular}[c]{@{}c@{}}Batch\\Size\end{tabular} & \begin{tabular}[c]{@{}c@{}c@{}}Gradient\\Accu. Steps\end{tabular} & \begin{tabular}[c]{@{}c@{}}Warmup\\Portion\end{tabular}& GPUs \\\midrule
XL & 4 & 10 & 8 & 0.2 & 4 \\
XXL & 4 & 2 & 4 & 0.2 & 4 \\
\bottomrule
\end{tabular}
\vspace{-1em}
\caption{The pre-training settings. ``Gradient Accu. Steps'' refers to the gradient accumulation steps.} 
\label{tab:pre_setting}
\end{table}

\begin{table}[t]
\centering\small
\begin{tabular}{@{\hspace{0.05cm}}l@{\hspace{0.05cm}}c@{\hspace{0.1cm}}c@{\hspace{0.1cm}}c@{\hspace{0.1cm}}c@{\hspace{0.1cm}}c@{\hspace{0.05cm}}}
\toprule
\begin{tabular}[c]{@{}c@{}}Model\\Scale\end{tabular} & Epoch & \begin{tabular}[c]{@{}c@{}}Batch\\Size\end{tabular} & \begin{tabular}[c]{@{}c@{}c@{}}Gradient\\Accu. Steps\end{tabular} & \begin{tabular}[c]{@{}c@{}}Warmup\\Portion\end{tabular}& GPUs \\\midrule
Brote-EX-XL & 4 & 10 & 8 & 0.2 & 4 \\
Brote-IM-XL & 2 & 10 & 8 & 0.2 & 4 \\\midrule
Brote-EX-XXL & 2 & 2 & 4 & 0.2 & 4 \\
Brote-IM-XXL & 1 & 1 & 4 & 0.2 & 4 \\
\bottomrule
\end{tabular}
\vspace{-1em}
\caption{The fine-tuning settings. ``Gradient Accu. Steps'' refers to the gradient accumulation steps.} 
\label{tab:fine_setting}
\vspace{-1em}
\end{table}

\section{Baselines}\label{appendix:baseline}
We compare to MLLMs who also notice the multi-image scenarios, including MMICL~\cite{zhao2023mmicl}, Otter~\cite{li2023otter}, VPG-C~\cite{li2023fine}, KOSMOS-1~\cite{huang2023language} and EMU~\cite{sun2023generative}. For models that are not initially designed for accepting multiple images, such as InstructBLIP, we concatenate the visual embeddings for all the input image together to enable the multi-image processing ability. The details of the baselines are listed together with the results in Table~\ref{tab:main_multi_full} and Table~\ref{tab:main_single_full}. 

\begin{table*}[!t]
\centering\small
\begin{tabular}{lccccc}
\toprule
Benchmark & Data Format & Answer Type & Setting & Data Split & Metrics \\ \midrule
NLVR2 & Multi-image & True/False & Zero-shot & Test & Accuracy \\
DEMON-Core & Multi-image & \begin{tabular}[c]{@{}c@{}}Multiple-choice\\ \& Open-ended\end{tabular} & Zero-shot & - &  I4-score \\
\noalign{\vskip 1pt} \cdashline{1-6} \noalign{\vskip 2pt}
MSVD QA & Video & Open-ended & Zero-shot & Test &  Accuracy \\
MSRVTT QA & Video & Open-ended & Zero-shot & Test &  Accuracy \\
SEED Bench & \begin{tabular}[c]{@{}c@{}}Video \&\\ Single image \end{tabular} & Multiple-choice & Zero-shot & - &  Accuracy \\
\noalign{\vskip 1pt} \cdashline{1-6} \noalign{\vskip 2pt}
VQAv2 & Single-image & Open-ended & Zero- \& few-shot & Test  & Soft accuracy\\
A-OKVQA & Single-image & Open-ended &  Zero- \& few-shot & Val  & Soft accuracy\\
ScienceQA-IMG & Single-image & Multiple-choice & Zero-shot CoT & Test  & Accuracy\\
MMBench & Single-image & Multiple-choice & Zero-shot & Dev &  Accuracy+\\
MME & Single-image & Yes/No & Zero-shot & - &  Accuracy+\\
\bottomrule
\end{tabular}
\caption{An overview evaluation benchmarks and metrics.} 
\label{tab:eval_bench}
\end{table*}

\section{Benchmarks and Metrics for Evaluation} \label{appendix:metrics}
The employed benchmarks and corresponding metrics are listed in Table~\ref{tab:eval_bench}. We investigate diverse conventional VL benchmarks and recently proposed MLLM benchmarks, including VQAv2~\cite{goyal2017making}, A-OKVQA~\cite{schwenk2022aok}, ScienceQA~\cite{saikh2022scienceqa}, NLVR2~\cite{suhr2019corpus}, MSVD QA~\cite{xu2017video}, MSRVTT QA~\cite{xu2017video}, SEED-Bench~\cite{li2023seedbench}), DEMON~\cite{li2023fine}), 
MME~\cite{fu2023mme} and MMBench~\cite{liu2023mmbench}.

For VQAv2 and A-OKVQA, we test the zero-shot question answering ability given a single image, and also evaluate the ability to gain information from other related images under few-shot ICL setting. ScienceQA is proposed for Chain-of-Thought (CoT)~\cite{wei2022chain} scenario, and we adopt the corresponding zero-shot CoT setting. SEED-Bench~\cite{li2023seedbench} is a recently proposed benchmark that also aims at question answering, which comprises both images and videos. We evaluate the zero-shot ability on it because no training set is available for extracting few-shot examples. As NLVR2 contains naturally interleaved image-text instances, we only conduct zero-shot evaluation. With recently proposed benchmarks for MLLMs, such as MME~\cite{fu2023mme}, MMBench~\cite{liu2023mmbench}, and DEMON~\cite{li2023fine}, we employ the zero-shot setting. 

Following previous works~\cite{antol2015vqa, saikh2022scienceqa, suhr2019corpus, huang2023language}, we report the soft-accuracy scores~\cite{antol2015vqa} for A-OKVQA and VQAv2, and calculate the accuracy scores on ScienceQA, NLVR2, MSVD QA and MSRVTT QA. The accuracy+ is employed as the metric for MME, and I4-score~\cite{li2023fine} is used for DEMON-Core. 

\begin{table*}[t]
\centering
\footnotesize
\begin{tabular}{lc@{\hspace{0.2cm}}c|c@{\hspace{0.15cm}}cc@{\hspace{0.2cm}}c@{\hspace{0.2cm}}c@{\hspace{0.2cm}}c@{\hspace{0.2cm}}c}
\toprule 
\multirow{2}{*}[-2ex]{Model} & \multirow{2}{*}[-2ex]{\begin{tabular}[c]{@{}c@{}}LLM\\Backbone\end{tabular}} & \multirow{2}{*}[-2ex]{\begin{tabular}[c]{@{}c@{}}\#Param\\LLM\end{tabular}} & \multicolumn{2}{c}{In-context Learning} & \multicolumn{5}{c} {Multi-image / Video Tasks} \\ \cmidrule(lr){4-5}\cmidrule(lr){6-10} 
 & & & VQAv2 & A-OKVQA & NLVR2  & DEMON & SEED & \begin{tabular}[c]{@{}c@{}}MSVD\\QA\end{tabular} & \begin{tabular}[c]{@{}c@{}}MSRVTT\\QA\end{tabular} \\
\midrule \noalign{\vskip -3pt}
\multicolumn{10}{c}{\cellcolor[HTML]{EFEFEF} \scriptsize\textit{Models SMALLER than 10B}}    \\ \noalign{\vskip 1pt}
KOSMOS-1 & MAGNETO & 1.3B & 51.80\hphantom{$^*$} & - & - & - & - & - & - \\
KOSMOS-2 & MAGNETO & 1.3B & - & - & - & - & 50.00\hphantom{$^*$} & - & -  \\
InstructBLIP-XL & FlanT5 & 3B & 31.76$^*$ & 39.13$^*$ & 52.59$^*$  & 32.59$^*$ & 52.70\hphantom{$^*$} & 43.40$^*$ & 12.12$^*$  \\  
MMICL-XL$^{\diamondsuit}$  & FlanT5 &  3B  & 69.16\hphantom{$^*$} &  53.43$^*$ & \underline{71.48}$^*$  & \textbf{38.14}$^*$ & 54.69$^*$ & \underline{53.68}\hphantom{$^*$} & 42.36$^*$  \\ 
Otter    & MPT &   7B   & 45.39$^*$ & 38.42$^*$ & 49.54$^*$  & 24.51\hphantom{$^*$} & 39.70\hphantom{$^*$} & 25.87$^*$ & \hphantom{0}9.78$^*$\\
VPG-C-LLaMA2  & LLaMA &  7B & - & 34.29$^*$ & 53.82$^*$ & 37.22\hphantom{$^*$} & - & \hphantom{0}6.03$^*$ & - \\ 
Flamingo-9B & Chinchilla & 7B & 56.3\hphantom{0$^*$} & - & - & - & - & 30.2\hphantom{0$^*$} & 13.7\hphantom{0$^*$} \\ 
\noalign{\vskip 1pt} \cdashline{1-10} \noalign{\vskip 2pt}
Brote-EX-XL(ours)    & FlanT5 &  3B  & \textbf{69.97}\hphantom{$^*$}  & \underline{56.00}\hphantom{$^*$}  & 71.41\hphantom{$^*$}  & 37.33\hphantom{$^*$} & \underline{57.51}\hphantom{$^*$} & 53.02\hphantom{$^*$} & \underline{43.14}\hphantom{$^*$} \\
Brote-IM-XL(ours)    & FlanT5 &  3B  &\underline{68.94}\hphantom{$^*$} &  \textbf{56.43}\hphantom{$^*$}  & \textbf{76.02}\hphantom{$^*$} & \underline{37.34}\hphantom{$^*$} & \textbf{57.86}\hphantom{$^*$} & \textbf{56.06}\hphantom{$^*$} & \textbf{45.08}\hphantom{$^*$} \\
\midrule \noalign{\vskip -3pt}
\multicolumn{10}{c}{\cellcolor[HTML]{EFEFEF} \scriptsize\textit{Models LARGER than 10B}}    \\ \noalign{\vskip 1pt}
InstructBLIP-XXL & FlanT5 &  11B    & 48.21$^*$  & 45.92$^*$  & 64.54$^*$    & 33.00$^*$  & 50.81$^*$  & 44.30$^*$ &15.49$^*$  \\  
MMICL-XXL$^{\diamondsuit}$   & FlanT5 &  11B    & 70.56\hphantom{$^*$}  & 54.85$^*$  & 56.16$^*$   & 36.30$^*$  & 56.66$^*$  & 52.19\hphantom{$^*$} & 39.46$^*$  \\
VPG-C-Vicuna  & Vicuna & 13B & - & - & - & 36.37\hphantom{$^*$} & - & - & - \\ 
BLIP-2-13B &  Vicuna & 13B & - & - & - & - & 46.4\hphantom{0$^*$} & 20.3\hphantom{0$^*$} & 10.3\hphantom{0$^*$} \\ 
InstructBLIP-13B & Vicuna & 13B & - & - & - & - & - & 41.2\hphantom{0$^*$} & 24.8\hphantom{0$^*$} \\ 
EMU-I   & LLaMA &  13B    & 58.4\hphantom{0$^*$} & - & - & - & - & 37.0\hphantom{0$^*$} & 21.2\hphantom{0$^*$}  \\
EMU-2   & LLaMA &  33B    & 67.0\hphantom{0$^*$} & - & - & - & \textbf{62.8}\hphantom{0$^*$} & 49.0\hphantom{0$^*$} & 31.4\hphantom{0$^*$}  \\
Flamingo-80B & Chinchilla & 70B & 63.1\hphantom{0$^*$} & - & - & - & - & 35.6\hphantom{0$^*$} & 17.4\hphantom{0$^*$} \\ 
\noalign{\vskip 1pt} \cdashline{1-10} \noalign{\vskip 2pt}
Brote-EX-XXL(ours)   &  FlanT5 &  11B  & \underline{70.86}\hphantom{$^*$} & \underline{59.94}\hphantom{$^*$}  & \underline{70.42}\hphantom{$^*$} & \underline{38.70}\hphantom{$^*$}& 59.31\hphantom{$^*$} & \underline{54.52}\hphantom{$^*$} & \underline{45.24}\hphantom{$^*$} \\ 
Brote-IM-XXL(ours) & FlanT5  &  11B  &\textbf{71.71}\hphantom{$^*$} &\textbf{60.31}\hphantom{$^*$} & \textbf{80.71}\hphantom{$^*$} &\textbf{38.94}\hphantom{$^*$} &\underline{61.64}\hphantom{$^*$} &\textbf{57.29}\hphantom{$^*$} & \textbf{45.94}\hphantom{$^*$} \\ 
\bottomrule
\end{tabular}
\caption{
Results for multi-image settings. The best results for models larger/smaller than 10B are separately \textbf{bolded} and the seconds are \underline{underlined}. $^{\diamondsuit}$: the InstructBLIP version. We evaluate results which are not officially announced using public checkpoints and mark them by *. SEED refers to SEED-Bench that contains both images and videos.} 
\label{tab:main_multi_full}
\end{table*}

\begin{table*}[t]
\centering\footnotesize
\begin{tabular}{@{\hspace{0.05cm}}lc@{\hspace{0.2cm}}c|c@{\hspace{0.2cm}}c@{\hspace{0.2cm}}c@{\hspace{0.2cm}}|c@{\hspace{0.2cm}}c@{\hspace{0.2cm}}c}
\toprule 
Model & \begin{tabular}[c]{@{}c@{}}LLM\\Backbone\end{tabular} & \begin{tabular}[c]{@{}c@{}}\#Param\\LLM\end{tabular} & \begin{tabular}[c]{@{}c@{}}VQAv2\\ \scriptsize{0-shot}\end{tabular} & \begin{tabular}[c]{@{}c@{}}A-OKVQA\\ \scriptsize{0-shot}\end{tabular} & \begin{tabular}[c]{@{}c@{}}ScienceQA\\ \scriptsize{-IMG}\end{tabular} & \begin{tabular}[c]{@{}c@{}}MME\\ \scriptsize{Perception}\end{tabular}  & \begin{tabular}[c]{@{}c@{}}MME\\ \scriptsize{Cognition}\end{tabular} & MMBench \\ \midrule \noalign{\vskip -3pt}
  \multicolumn{9}{c}{\cellcolor[HTML]{EFEFEF} \scriptsize\textit{Models SMALLER than 10B}}    \\ \noalign{\vskip 1pt}
KOSMOS-1 & MAGNETO &1.3B & 51.80\hphantom{$^*$} & - & - & - & - & -  \\
InstructBLIP-XL & FlanT5  &3B    & 36.77\hphantom{$^*$} & \textbf{54.57}\hphantom{$^*$} & 70.40\hphantom{$^*$} & 1093.70$^*$ & 281.43$^*$ & 69.68$^*$  \\  
MMICL-XL      & FlanT5  &3B  & 69.13\hphantom{$^*$}  &  52.12$^*$  & \textbf{72.58}$^*$  & 1184.54$^*$  & 277.86$^*$   & 73.11$^*$    \\ 
LLaVA$^\dag$  & LLaMA & 7B &  - & - & - & 457.82 & 214.64\hphantom{$^*$} & 36.2\hphantom{0$^*$} \\
Otter$^\dag$      & MPT & 7B   & 57.89$^*$ & 41.92$^*$  & 63.10\hphantom{$^*$}  & \underline{1292.26}\hphantom{$^*$} & \underline{306.43}\hphantom{$^*$} & 48.3\hphantom{0$^*$} \\
VPG-C-Vicuna  & Vicuna & 7B   & - & - & - & \textbf{1299.24}\hphantom{$^*$} & \textbf{321.07}\hphantom{$^*$} & - \\
\noalign{\vskip 1pt} \cdashline{1-9} \noalign{\vskip 2pt}
Brote-EX-XL(ours)   & FlanT5 & 3B  & \underline{69.90}\hphantom{$^*$}  & 52.93\hphantom{$^*$}  & \underline{71.15}\hphantom{$^*$} & 1203.87\hphantom{$^*$} & 301.79\hphantom{$^*$} & \underline{73.27}\hphantom{$^*$}  \\
Brote-IM-XL(ours)   & FlanT5 & 3B  & \textbf{70.24}\hphantom{$^*$} &  \underline{53.40}\hphantom{$^*$}  & \textbf{72.58}\hphantom{$^*$}  & 1181.95\hphantom{$^*$} & 266.79\hphantom{$^*$} & \textbf{74.29}\hphantom{$^*$}  \\
\midrule \noalign{\vskip -3pt}
  \multicolumn{9}{c}{\cellcolor[HTML]{EFEFEF} \scriptsize\textit{Models LARGER than 10B}}    \\ \noalign{\vskip 1pt}
InstructBLIP-XXL & FlanT5 & 11B    & 63.69 & \underline{57.10}\hphantom{$^*$} & 70.60\hphantom{$^*$} & 1212.82$^*$  & 291.79$^*$  & 70.34$^*$    \\  
JiuTian$^\dag$      & FlanT5 &  11B    & -  & - & - & - & - & 64.7\hphantom{0$^*$}  \\
MMICL-XXL      & FlanT5 &  11B    & 70.30  & 51.35$^*$  & 74.92$^*$ & \underline{1313.88}$^*$  & \underline{311.79}$^*$  & 76.58$^*$   \\
MMICL-XXL (BLIP2)$^\dag$     & FlanT5 &  11B    &69.99 &- & -& \textbf{1381.74}\hphantom{$^*$} & \textbf{428.93}\hphantom{$^*$} & 65.24\hphantom{$^*$}   \\
\noalign{\vskip 1pt} \cdashline{1-9} \noalign{\vskip 2pt}
Brote-EX-XXL(ours)   & FlanT5 &  11B  & \underline{71.58} & 56.47\hphantom{$^*$}  & \underline{77.69}\hphantom{$^*$} & 1279.73\hphantom{$^*$} &310.01\hphantom{$^*$} &\underline{76.67}\hphantom{$^*$}  \\ 
Brote-IM-XXL(ours)   & FlanT5 &  11B  &\textbf{73.02} & \textbf{57.83}\hphantom{$^*$} &\textbf{78.38}\hphantom{$^*$} &1284.13\hphantom{$^*$} &300.00\hphantom{$^*$} & \textbf{77.34}\hphantom{$^*$} \\ 
\bottomrule
\end{tabular}
\caption{Zero-shot results for single-image settings. The best results for models larger/smaller than 10B are separately \textbf{bolded} and the seconds are \underline{underlined}. 
$^\dag$: results of these models are taken from \citet{zhao2023mmicl}. We evaluate results which are not officially announced using public checkpoints and mark them by *.} 
\label{tab:main_single_full}
\end{table*}

\begin{table*}[t]
\centering\small
\begin{tabular}{@{\hspace{0.1cm}}l@{\hspace{0.05cm}}c@{\hspace{0.15cm}}|@{\hspace{0.15cm}}c@{\hspace{0.1cm}}c@{\hspace{0.1cm}}c@{\hspace{0.1cm}}c@{\hspace{0.1cm}}c@{\hspace{0.1cm}}c@{\hspace{0.1cm}}c@{\hspace{0.1cm}}|@{\hspace{0.15cm}}c@{\hspace{0.05cm}}}
\toprule
 Model & \begin{tabular}[c]{@{}c@{}}\#Param\\LLM\end{tabular} & \begin{tabular}[c]{@{}c@{}}Multimodal\\Dialogue\end{tabular} & \begin{tabular}[c]{@{}c@{}}Visual\\ Storytelling\end{tabular} & \begin{tabular}[c]{@{}c@{}}Visual Rel. \\Inference\end{tabular} & \begin{tabular}[c]{@{}c@{}}Multimodal\\Cloze\end{tabular} & \begin{tabular}[c]{@{}c@{}}Knowledge\\ QA\end{tabular} & \begin{tabular}[c]{@{}c@{}}Text-rich\\Images QA\end{tabular} & \begin{tabular}[c]{@{}c@{}}Multi-image\\ Reasoning\end{tabular}   & AVG \\  \midrule 
MiniGPT-4$^\dag$  &  7B & 13.69 & 17.07 & 7.95 & 16.60 & 30.27 & 26.40 &  43.50 & 22.21 \\
Otter$^\dag$  &  7B  &  15.37 &  15.57 & 11.39   &  16.00   &  41.67  &  27.73  &  43.85  &  24.51  \\
BLIP-2-XXL$^\dag$  &  11B  &  26.12 & 21.31 & 10.67 & 17.94 & 39.23 & 33.53 & 39.65 &  26.92   \\
InstructBLIP-XL$^\diamondsuit$  &  3B  & 19.42	& 25.09	 & 15.21 & 32.35 & 48.13 & 38.89 & 49.04 & 32.59 \\
InstructBLIP-XXL$^\dag$  &  11B  &  33.58  &  24.41  &  11.49 &  21.20  &  47.40  & 44.40  &  48.55  &  33.00   \\
MMICL-XXL$^\diamondsuit$   &  11B  &  31.60  &  \textbf{28.76}  &  12.17 & 31.86 &  \underline{61.58} &  44.33 &  43.73 &  36.30  \\
VPG-C-Vicuna$^\dag$  &  13B & \underline{37.50} & 25.20 & \textbf{25.90} & 22.15 & 48.60 & \underline{49.93} & 50.28 & 36.37 \\ 
VPG-C-LLaMA2$^\dag$  &  7B & \textbf{42.70} & 24.76 & \underline{25.50} & 22.95 & 51.00 & 44.93 & 48.68 & 37.22 \\ 
MMICL-XL$^\diamondsuit$   &   3B  & 33.32 & 27.14 & 13.58 & \textbf{34.17} & 58.45 & 47.19 & \textbf{53.10} & 38.14  \\
\noalign{\vskip 1pt} \cdashline{1-10}\noalign{\vskip 2pt}
Brote-XL    &   3B  & 32.46 & 27.38 & 10.51 & \underline{32.41} & 59.45 & 48.07  & 51.08 & 37.34 \\ 
Brote-XXL    &  11B   &  34.95 & \underline{28.23} & 11.11 & 29.51 & \textbf{65.25} & \textbf{50.87}  & \underline{52.65} & \textbf{38.94} \\ 
\bottomrule
\end{tabular}
\caption{Evaluation on DEMON-Core benchmark. Models marked by $^\dag$: results taken from \citet{li2023fine}. Models marked by $^\diamondsuit$: we evaluate the results with official checkpoints as stated in Appendix~\ref{appendix:full_rst}.} 
\label{tab:demon}
\end{table*}

\section{Full results of other popular MLLMs} \label{appendix:full_rst}
As we evalution the performance on a variaty of task, where some results are missing for certain closel related baselines. We use the public checkpoints to obtain the missing results for MMICL, InstructBLIP and Otter \footnote{The detailed model versions with links are as follows:\\ MMICL (\url{https://huggingface.co/BleachNick/MMICL-Instructblip-T5-xl} and \url{https://huggingface.co/BleachNick/MMICL-Instructblip-T5-xxl});\\ InstructBLIP (\url{https://huggingface.co/Salesforce/instructblip-flan-t5-xl} and \url{https://huggingface.co/Salesforce/instructblip-flan-t5-xxl});\\Otter: (\url{https://huggingface.co/luodian/OTTER-Image-MPT7B}).}, together with official scripts and required environments. 
Apart from the MLLMs that focus on interleaved and instruction-following settings, we also provide a table of results from other popular MLLMs. Table~\ref{tab:main_multi_full} and Table~\ref{tab:main_single_full} record the results for multi-image and single-image settings respectively. The detailed results of the subtasks from DEMON-core are listed in Table~\ref{tab:demon}.

In Table~\ref{tab:main_multi_full} and Table~\ref{tab:demon}, our models demonstrate better performance over the others of different scales, including models of larger scales. We outperform strong baselines, such as InstructBLIP, MMICL and VPG-C, which include also consider prior-LLM instruction-image fusion. This support our finding that our browse-and-concentrate paradigm contributes to a more in-depth understanding of multimodal context with the assistant of these intermediate browsing insights. 

However, for single-image tasks reported in Table~\ref{tab:main_single_full}, we notice a different trend on MME benchmark. For models with LLMs small than 10B, VPG-C-Vicuna and Otter show impressive performance on MME. For models with LLMs larger than 10B, MMICL-XXL (BLIP2) presents the best performance, followed by its variant MMICL-XXL (InstructBLIP). Our models only outperform InstructBLIP models. This is potentially caused by the limitaion of training data, where we exclude the visual instruction tuning dataset such as LLaVA~\cite{liu2023visual} during pre-training and fine-tuning, because the outputs can vary subjectively. On the contrary, our models continue to manifest progress for single-image QA tasks and the other MLLM benchmark MMBench.

\section{Details for Ablation Study on the Training Strategies} \label{appendix:abla}
In this section, we provide the detailed results for ablation study of our proposed strategies as an accompany of Table~\ref{tab:abla_strategy}. Table~\ref{tab:abla_strategy_full} lists the results for each tasks, and averaged scores for multi-image tasks (AVG-Multi), single-image tasks (AVG-Single) and overall tasks (AVG).
In the settings without context-dropping strategies,, our model with pre-training presents superior multi-image comprehension, as evidenced by its performance on the A-OKVQA 4-shot and SEED-video settings, in comparison to its counterpart without pre-training. Nonetheless, without context-dropping strategies, both models exhibit a limitation in achieving a balanced performance across single-image and multi-image scenarios. To address this, we incorporate context-dropping strategies designed to encourage the models to effectively utilize the given condition context vector, as detailed in Section~\ref{sec:fine_strategy}. We eventually adopt the ``Drop-ALL'' setting for training  our Brote models.

\begin{table*}[t]
\centering\small
\begin{tabular} {@{\hspace{0.1cm}}c@{\hspace{0.1cm}}c@{\hspace{0.1cm}}c@{\hspace{0.1cm}}c@{\hspace{0.1cm}}|c@{\hspace{0.2cm}}cc@{\hspace{0.2cm}}c@{\hspace{0.15cm}}c@{\hspace{0.15cm}}c@{\hspace{0.15cm}}c@{\hspace{0.15cm}}|@{\hspace{0.15cm}}c@{\hspace{0.15cm}}c@{\hspace{0.15cm}}c@{\hspace{0.1cm}}}\toprule
\multirow{2}{*}[-0.8ex]{Models}  & \multirow{2}{*}[-0.8ex]{\begin{tabular}[c]{@{}c@{}}Pre\\-train\end{tabular}} & \multirow{2}{*}[-0.8ex]{\begin{tabular}[c]{@{}c@{}}Fine\\-tune\end{tabular}}  &  \multirow{2}{*}[-0.8ex]{Drop}  & \multicolumn{2}{c}{A-OKVQA} & \multicolumn{2}{c}{SEED} & \multirow{2}{*}[-1ex]{NLVR2} & \multirow{2}{*}[-1ex]{\begin{tabular}[c]{@{}c@{}}SciQA\\-IMG\end{tabular}} & \multirow{2}{*}[-1ex]{MSVD} & \multirow{2}{*}[-1ex]{\begin{tabular}[c]{@{}c@{}}AVG\\-Multi\end{tabular}} & \multirow{2}{*}[-1ex]{\begin{tabular}[c]{@{}c@{}}AVG\\-Single\end{tabular}} & \multirow{2}{*}[-1ex]{AVG} \\  \cmidrule(lr){5-6}\cmidrule(lr){7-8}
    &  &  &  & 0shot  & 4shot & Image  & Video  &  &  &  &     \\ \midrule
\textcolor{gray}{InstructBLIP}  & \textcolor{gray}{-} & \textcolor{gray}{-} & \textcolor{gray}{-}  & \textcolor{gray}{54.57} & \textcolor{gray}{39.13} & \textcolor{gray}{-} & \textcolor{gray}{-} & \textcolor{gray}{52.59} & \textcolor{gray}{70.40} & \textcolor{gray}{43.40} & \textcolor{gray}{47.05} & \textcolor{gray}{57.85} & \textcolor{gray}{51.68}\\ \midrule
\textcolor{gray}{MMICL}  & \textcolor{gray}{-} & \textcolor{gray}{-} & \textcolor{gray}{-}  & \textcolor{gray}{51.53} & \textcolor{gray}{53.32} & \textcolor{gray}{58.81} & \textcolor{gray}{35.40} & \textcolor{gray}{62.40} & \textcolor{gray}{63.21} & \textcolor{gray}{37.07} & \textcolor{gray}{47.05} & \textcolor{gray}{57.85} & \textcolor{gray}{51.68}\\ \midrule
Ours-sampled  & \ding{56}  & \checkmark & \ding{56}  & \textbf{53.15} & 54.76 & \textbf{60.38} & 33.76 & 63.35 & 63.36 & 35.87 & 46.51 & \textbf{58.96} & 51.85  \\
Ours-sampled   & \checkmark & \checkmark & \ding{56}  & 49.94 & 56.16 & 57.57 & 36.85 & 60.58 & 60.58 & 34.17 &  47.12 & 56.03 & 50.94 \\
\noalign{\vskip 1pt} \cdashline{1-14}\noalign{\vskip 2pt}
Ours-sampled   & \checkmark & \checkmark & IMG-N & 50.39 & 54.79 & 60.16 & 37.31 & 64.62 & 61.87 & 35.51 & 48.06 & 57.47 & 52.09   \\
Ours-sampled   & \checkmark & \checkmark & IMG-B & 48.53 & 55.22 & 59.21 & 37.57 & 65.00 & \textbf{63.31} & 34.36 & 48.06 & 57.02 & 51.90  \\
Ours-sampled   & \checkmark & \checkmark & TXT & 50.14 & 54.51 & 59.14 & 36.90 & 61.56 & 62.92 & \textbf{39.36} & 48.08 & 57.40 & 52.08  \\ 
Ours-sampled   & \checkmark & \checkmark & All & 48.22 & \textbf{55.35} & 59.86 & \textbf{37.64} & \textbf{65.15} & 63.16 & 37.35 & \textbf{48.87} & 57.08 & \textbf{52.39}  \\ 
\bottomrule
\end{tabular}
\caption{Ablation study of different training strategies on XL-sized (3B LLM) models trained with sampled subset, where ``Ours'' refers to \textit{Our-sampled} described in \S\ref{sec:abla_strategy}. ``AVG-Multi'' is the averaged over A-OKVQA 4-shot, SEED image split, NLVR2 and MSVD, and ``AVG-Single'' is the averaged over the rest. ``AVG'' refers to the average accuracy of all the tasks in this table.} 
\label{tab:abla_strategy_full}
\end{table*}

\end{document}